\documentclass{article}

\usepackage[final]{neurips_2025}

\usepackage{amsmath}
\usepackage{amssymb}
\usepackage{mathtools}
\usepackage{amsthm}
\usepackage{xspace}

\usepackage{natbib}
\PassOptionsToPackage{table}{xcolor}
\usepackage{url}
\usepackage{booktabs}
\usepackage{siunitx}
\usepackage[ruled,vlined]{algorithm2e}
\usepackage{algorithmic}
\usepackage{tikz}
\usetikzlibrary{tikzmark,calc}
\usepackage{multirow}
\usepackage{xcolor}
\usepackage{subcaption}
\usepackage{svg}
\usepackage{colortbl}
\usepackage{pifont} 
\usepackage{makecell}
\definecolor{lightblue}{rgb}{0.1, 0.1, 0.9} 
\usepackage{fancyhdr}

\theoremstyle{plain}
\theoremstyle{definition}

\theoremstyle{remark}

\newcommand{\our}{\texttt{SageAttention3}\xspace}
\newcommand{\sageback}{\texttt{SageBwd}\xspace}
\newcommand{\jt}[1]{\textcolor{blue}{{#1}}\xspace}

\newcommand{\rowmax}{\mathrm{rowmax}}

\newcommand{\mean}{\mathrm{mean}}
\newcommand{\annotate}[1]{\textcolor{gray}{{#1}}\xspace}
\newcommand{\cogvideo}{\texttt{CogvideoX}\xspace}
\newcommand{\hyvideo}{\texttt{HunyuanVideo}\xspace}
\newcommand{\mochi}{\texttt{Mochi}\xspace}

\newcommand{\qwen}{\texttt{Qwen2.5}\xspace}
\newcommand{\llamal}{\texttt{Llama3.2}\xspace}

\newcommand{\flux}{\texttt{Flux}\xspace}
\newcommand{\sd}{\texttt{Stable-Diffusion3.5}\xspace}
\usepackage{float} 
\usepackage{xcolor}
\usepackage{placeins} 
\usepackage[utf8]{inputenc} % allow utf-8 input
\usepackage[T1]{fontenc}    % use 8-bit T1 fonts
\usepackage{hyperref}       % hyperlinks
\usepackage[capitalize,noabbrev]{cleveref}
\usepackage{url}            % simple URL typesetting
\usepackage{booktabs}       % professional-quality tables
\usepackage{diagbox}
\usepackage{lscape}
\usepackage{amsfonts}       % blackboard math symbols
\usepackage{nicefrac}       % compact symbols for 1/2, etc.
\usepackage{microtype}      % microtypography
\usepackage{xcolor}         % colors

\title{SageAttention3: Microscaling FP4 Attention for Inference and An Exploration of 8-bit Training}

\author{%
Jintao Zhang$^{*12}$, 
Jia Wei$^{*1}$, 
Haoxu Wang$^{1}$, 
Pengle Zhang$^{1}$, 
Xiaoming Xu$^{1}$, 
Haofeng Huang$^{1}$,\\
\textbf{Kai Jiang}$^{1}$, 
\textbf{Jianfei Chen}$^{\dagger1}$, 
\textbf{Jun Zhu}$^{\dagger12}$\\[1mm]
$^{1}$Dept. of Comp. Sci. and Tech., Institute for AI, BNRist Center, THBI Lab,\\
Tsinghua-Bosch Joint ML Center, Tsinghua University; \quad $^{2}$Shengshu Tech., Beijing, China.\\[1mm]
\texttt{\{zhang-jt24@mails., jianfeic@, dcszj@\}tsinghua.edu.cn}
}

\newcommand{\ourf}{\texttt{SageAttention3}\xspace}
\NewDocumentCommand{\jintao}{ mO{} }{\textcolor{blue}{\textsuperscript{\textit{JT}}\textsf{\small[#1]}}}

\definecolor{deepgreen}{rgb}{0.0, 0.5, 0.0}  
\definecolor{deepred}{rgb}{0.6, 0.0, 0.0}

\newcommand{\vQ}{\mathbf{Q}}
\newcommand{\vK}{\mathbf{K}}
\newcommand{\vV}{\mathbf{V}}
\newcommand{\vdQ}{\mathbf{dQ}}
\newcommand{\vdK}{\mathbf{dK}}
\newcommand{\vdV}{\mathbf{dV}}
\newcommand{\vS}{\mathbf{S}}
\newcommand{\vdS}{\mathbf{dS}}
\newcommand{\vP}{\mathbf{P}}
\newcommand{\vdP}{\mathbf{dP}}

\newcommand{\vO}{\mathbf{O}}
\newcommand{\vdO}{\mathbf{dO}}

\begin{document}

\maketitle

{\renewcommand{\thefootnote}{}\footnotetext{%
$^{*}$ co-first authors, $^{\dagger}$ Corresponding authors.}}

\begin{abstract}
The efficiency of attention is important due to its quadratic time complexity. We enhance the efficiency of attention through two key contributions: 
First, we leverage the new \texttt{FP4} Tensor Cores in Blackwell GPUs to accelerate attention computation. Our implementation achieves \textbf{1038} \texttt{TOPS} on \texttt{RTX5090}, which is a \textbf{5}$\times$ speedup over the fastest FlashAttention on \texttt{RTX5090}. Experiments show that our \texttt{FP4} attention can accelerate inference of various models in a plug-and-play way.
Second, we pioneer low-bit attention to training tasks. Existing low-bit attention works like FlashAttention3 and SageAttention focus only on inference. However, the efficiency of training large models is also important. To explore whether low-bit attention can be effectively applied to training tasks, we design an accurate and efficient \texttt{8-bit} attention for both forward and backward propagation. Experiments indicate that \texttt{8-bit} attention achieves lossless performance in fine-tuning tasks but exhibits slower convergence in pretraining tasks. The code is available at \url{https://github.com/thu-ml/SageAttention}.

\end{abstract}

\begin{figure}[ht]
    \centering
    \vspace{-.75em}
    \includegraphics[width=.91\columnwidth]{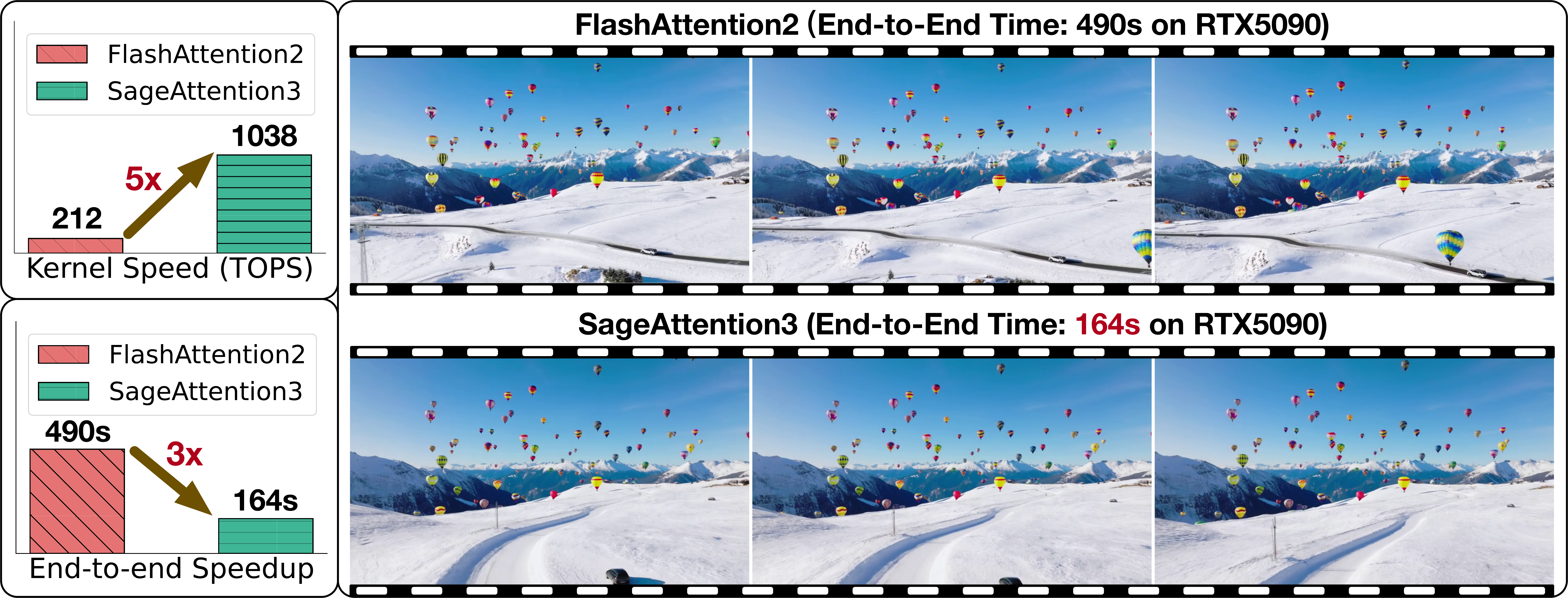} 
    % \vspace{-1em}
    \caption{The upper left figure shows the kernel speedup on \texttt{RTX5090}. The other two figures show the end-to-end inference speedup of generating a video using \texttt{HunyuanVideo} on \texttt{RTX5090}. Note that FlashAttention3 can only run on Hopper GPUs, so FlashAttention2 is already the fastest on \texttt{RTX5090}.}
    \vspace{-1em}
    \label{fig:cover} 
\end{figure}

\section{Introduction}
\textbf{Motivation}. 
The efficiency of attention is critical for generation models, especially given their quadratic time complexity with longer sequences~\citep{vaswani2017attention,zhangsurvey}. Quantization offers an effective way to accelerate inference by utilizing low-bit Tensor Cores in GPUs~\citep{2024sageattention,zhang2024sageattention2,zhang2025turbodiffusion,chen2020statistical}. The new \texttt{FP4} Tensor Cores in Blackwell GPUs deliver significantly faster performance compared to \texttt{FP16}~\citep{whitepaper5090}. We want to propose a novel \texttt{FP4} attention implementation that provides plug-and-play compatibility for inference acceleration. Beyond inference, training efficiency is equally important. However, no prior work has explored low-bit attention for training large models. To address this gap, we design a trainable \texttt{8-bit} attention to explore its feasibility in training tasks.

To the best of our knowledge, we are the \underline{first work} that designs \texttt{FP4} attention for inference and the \underline{first work} to explore the feasibility of low-bit attention for training large models.

\textbf{Challenges.} 
There are two primary obstacles for \texttt{FP4} attention and one key difficulty for \texttt{8-bit} trainable attention. First, \textbf{(C1)} \texttt{FP4} quantization suffers from severe value limitations (only 15 representable values), making both per-tensor and per-token quantization approaches inadequate for preserving model accuracy. Second, \textbf{(C2)} The attention map \( P \) consists primarily of small values in the range \([0, 1]\). When directly quantized to \texttt{FP4}, these values force the scaling factors into an extremely narrow dynamic range. However, hardware requires the quantization factors to be in \texttt{FP8} data type. This leads to significant accuracy loss when presenting these scale factors in \texttt{FP8}. 
Third, \textbf{(C3)} When employing \texttt{8-bit} attention during training, we find that the attention map gradients are particularly vulnerable to quantization errors, resulting in accumulated errors in the input gradients.

\textbf{Our Method.} To address \textbf{(C1)}, we propose to use \texttt{FP4} microscaling quantization for the two matrix multiplications in attention, i.e., $QK^\top$ and $PV$. 
By constraining the quantization group size to 1x16 (instead of per-tensor or per-channel), our method effectively contains outlier effects within each block while improving \texttt{FP4} quantization accuracy.
To overcome \textbf{(C2)}, we propose a two-level quantization method for $P$ to fully utilize the presentative range of the \texttt{FP8} scaling factor, enhancing the quantization accuracy of $P$. Specifically, this approach first normalizes each token's range to $\mathsf{[0, 448 \times 6]}$ through per-token quantization, then applies \texttt{FP4} microscaling quantization for enhanced precision.
To address \textbf{(C3)}, we identify the most accuracy-sensitive matrix multiplication among the five in backpropagation and maintain its accuracy in \texttt{FP16}.

\textbf{Result.} Our \texttt{FP4} attention, named \our, could achieve \textbf{1038} \texttt{TOPS} on \texttt{RTX5090}, which is a \textbf{5}$\times$ speedup than FlashAttention. Furthermore, we demonstrate that \texttt{8-bit} trainable attention, named \texttt{SageBwd}, could achieve lossless performance when fine-tuning base models for instruction-following tasks, but is not suitable for pretraining tasks. 

\textbf{Contribution.} Our work makes the following key contributions:

\textbf{(1)} We design the first \texttt{FP4} attention to accelerate inference, achieving \textbf{1000+} \texttt{TOPS} on \texttt{RTX5090}.

\textbf{(2)} We propose the first trainable low-bit attention, enabling accelerated training with lossless fine-tuning performance, while revealing key insights for low-bit attention in training.

\begin{figure}[ht]
    \centering
    \includegraphics[width=\columnwidth]{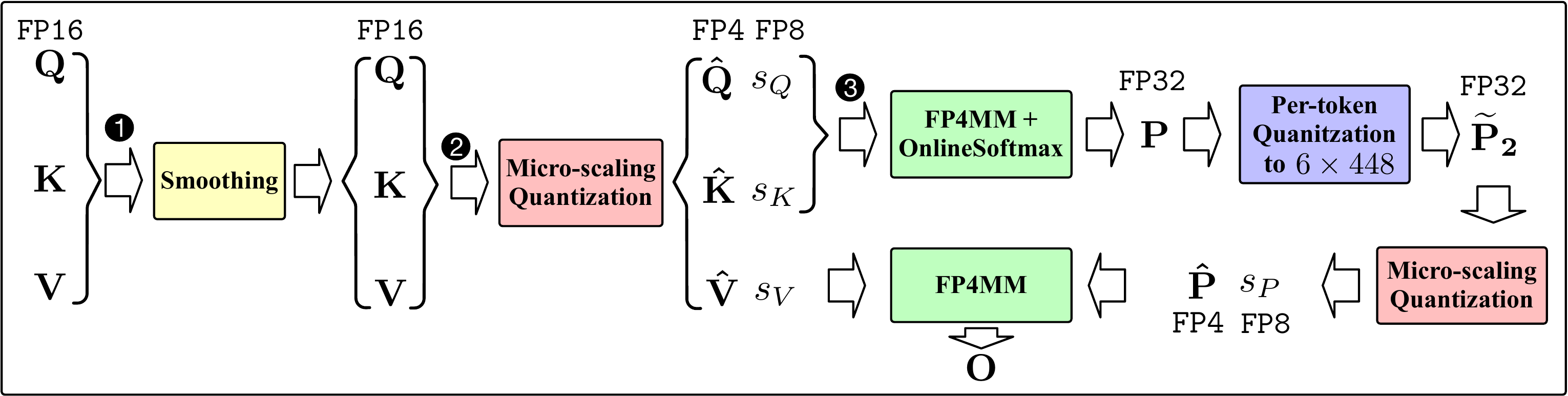} 
    \caption{Workflow of microscaling \texttt{FP4} attention.}
    \label{fig:workflow}
\end{figure}

\section{Preliminary}  \label{sec:preliminary}
\textbf{FlashAttention.} The attention computation contains two matrix multiplications and one softmax calculation: $S = QK^\top, P = \texttt{Softmax}(S), O = PV$. The $Q, K, V$ are in the shape of $N \times D$, where $N$ means the sequence length and $D$ means the dimension of an attention head. $P, S$ are in the shape of $N \times N$. FlashAttention divides $Q$ to blocks $\{Q_i\}$ in the shape of $B_q \times D$, and divides $K, V$ to $\{K_i\}, \{V_i\}$ in the shape of $B_{kv} \times D$. Then it uses online softmax to avoid the large memory IO for $S$ and $P$: $S_{ij} = Q_iK_j^\top, P_{ij} = \texttt{OnlineSoftmax}(S_{ij}), O_{ij} = P_{ij}V_{j}$.

\textbf{Notation.} For simplicity, we omit subscripts and use $\vQ, \vK, \vV, \vS, \vP, \vO$ to denote the matrix blocks in FlashAttention, while retaining full subscript notation in Algorithm~\ref{alg:sage3_fwd},~\ref{alg:int8_train_fwd}, and~\ref{alg:int8_train_bwd}.

\textbf{Quantization.} Quantization is used to accelerate Matmul by converting two matrices from high-bit to low-bit with scale factors. Take \texttt{INT8} quantization for Matmul $AB$ as an example, where $A$ and $B$ are in \texttt{FP16} data type. It can be formulated: $s_A = \max(|A|)/127, ~\hat A =  \lceil A / s_A \rfloor$, $s_B = \max(|B|)/127, ~\hat B = \lceil B / s_B \rfloor$, where $\hat A, \hat B$ are in \texttt{INT8} and the others are in \texttt{FP32}. Then, $AB \approx \hat A \hat B \times s_A \times s_B$, which can be accelerated by the \texttt{INT8} Tensor Core. The granularity of quantization is determined by the dimensions reduced by the $\max$ operation. For example, in \textit{per-token quantization}, the $\max$ is computed along each row of a matrix. In \textit{per-block quantization}, the $\max$ is computed on a block of a matrix, which in our paper means a FlashAttention block.

% \textbf{Quantization:}~   s_{ij} = \max(|X|)/6, ~~~   \hat X_{ij} =   \lceil X_{ij} / s_{ij} \rfloor,  \\
% \textbf{Dequantization:}~ [X_{ij}' = \phi^{-1}(\hat X_{ij}, s_{ij})]:~ X_{ij}' = s_{ij} \times \hat X_{ij}
% \label{equ:microscaling} 

\section{FP4 Attention for Inference Acceleration}
This section presents our microscaling \texttt{FP4} attention through three key components: (1) the fundamental workflow for applying microscaling \texttt{FP4} quantization to attention in Section~\ref{sec:micro_scaling_fp4}, (2) the two-level quantization approach for the attention map in Section~\ref{sec:two_level_quant}, and (3) critical hardware implementation optimization in Section~\ref{sec:hardware_implement}.

\begin{figure}[ht]
    \centering
    \includegraphics[width=1.01\columnwidth]{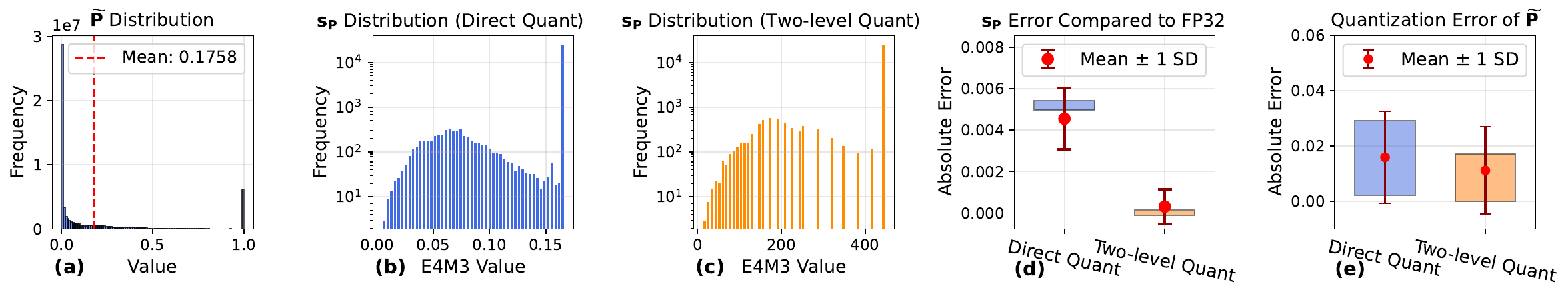} 
    \vspace{-.5em}
    \caption{Analysis of the benefit of two-level quantization. (a) shows the distribution of $\widetilde \vP$. (b) and (c) show the distribution of $\mathbf{s_P}$ using direct quantization and two-level quantization. (d) and (e) show the error of $\mathbf{s_P}$ and $\widetilde \vP$ using direct quantization and two-level quantization.}
    \label{fig:quant_errors_of_p} 
\end{figure}

\subsection{Microscaling {FP4} Attention}  \label{sec:micro_scaling_fp4}
% We formulate the Microscaling \texttt{FP4} Attention as follows.

\textbf{FP4 microscaling quantization}. Given a matrix $X \in \mathbb{R}^{N \times d}$, we quantize it to $\hat X$ in \texttt{FP4} data type with a scale factor matrix $s_X$ in \texttt{FP8} data type. Specifically, $X$ is partitioned into $X_{ij} \in \mathbb{R}^{1 \times n}$ blocks, where each $1 \times n$ block corresponds to one scale factor $s_{ij}$. The \texttt{FP4} microscaling quantization ($[\hat X, s_X = \phi(X)]$) and dequantization ($X' = \phi^{-1}(\hat X, s_X)$) can be formulated as follows.
\begin{align}
    \textbf{Quantization $\phi$:}~&   s_{ij} = \max(|X|)/6, ~~~   \hat X_{ij} =   \lceil X_{ij} / s_{ij} \rfloor  \\
    \textbf{Dequantization $\phi^{-1}$:}~& ~ X_{ij}' = s_{ij} \times \hat X_{ij}
    \label{equ:microscaling} 
\end{align}

Where the $\lceil \cdot \rfloor$ means \texttt{FP4} rounding.

\textbf{FP4 microscaling quantization Matmul}. Consider a matrix multiplication $AB$, where $A$ and $B$ are in \texttt{FP16} precision. The speed of the Matmul is about 200 \texttt{TOPS} on \texttt{RTX5090}. In contrast, the speed of the \texttt{FP4} microscaling Matmul is about 1600 \texttt{TOPS}, which is an 8x speedup. The \texttt{FP4} microscaling Matmul instruction (\texttt{FP4MM}) takes four inputs, i.e., $\hat A, s_A, \hat B, s_B$, and the output $C$ equals to the Matmul result between $\phi^{-1}(\hat A, s_A)$ and $\phi^{-1}(\hat B, s_B)$:
\begin{align}
    C = \texttt{FP4MM}(\hat A, s_A, \hat B, s_B)
    \label{equ:FP4MM} 
\end{align}
\textbf{Attention computation.} We accelerate attention computation by applying \texttt{FP4} microscaling quantization to both matrix multiplications: $\vQ\vK^\top$ and $\vP\vV$. 
\begin{align}
    \hat \vQ, \mathbf{s_Q} = \phi(\vQ), ~~~
    \hat \vK, \mathbf{s_K} =& \phi(\vK^\top), ~~~
    \vS = \texttt{FP4MM}(\hat \vQ, \mathbf{s_Q}, \hat \vK, \mathbf{s_K})  \notag \\
    \widetilde \vP =& \texttt{OnlineSoftmax}(\vS)  \notag \\
    \hat \vP, \mathbf{s_P} = \phi(\widetilde \vP), ~~~
    \hat \vV, \mathbf{s_V} =& \phi(\vV), ~~~~~~
    \vO = \texttt{FP4MM}(\hat \vP, \mathbf{s_P}, \hat \vV, \mathbf{s_V}) 
\end{align}

It is important to note that our hardware implementation builds on FlashAttention, where the matrices $\vQ$, $\vK$, $\widetilde \vP$, and $\vV$ in our formulation correspond to FlashAttention’s tiled $Q$, $K$, $\widetilde{P}$, and $V$ blocks as described in Section~\ref{sec:preliminary}. Additionally, to enhance the attention accuracy, we adopt the smoothing $Q$ and $K$ in SageAttention2~\citep{zhang2024sageattention2}. The complete algorithm is presented in Algorithm~\ref{alg:sage3_fwd}.

\textbf{Data type determination.} 
There are two choices for the \texttt{FP4} data type~\citep{rouhani2023microscaling}. The first one is the \texttt{NVFP4}, which is in \texttt{E2M1} data type and its quantization block size is $1 \times 16$ and its scale factor is in \texttt{E4M3} data type. The second one is the \texttt{MXFP4}, which is also in \texttt{E2M1} data type. However, its quantization block size is $1 \times 32$ and its scale factor is in \texttt{E8M0} data type. We choose \texttt{NVFP4} 
because the accuracy of \texttt{NVFP4} is much higher than that of \texttt{MXFP4} in attention quantization. \underline{Empirical results:} Table~\ref{tab:acc_ablation}(\subref{table:a}) shows the accuracy of \texttt{MXFP4} and \texttt{NVFP4} using real \textbf{Q, K, V} across all layers of \texttt{CogVideoX}. Results indicate that the accuracy of \texttt{NVFP4} outperforms that of \texttt{MXFP4}.

% \begin{table}[h]
% \caption{\textbf{Average accuracy} and \textbf{Worst accuracy} accross all layers of \texttt{CogVideoX} using different data types.}
% \centering
% \begin{minipage}{0.48\linewidth}
% \centering
% \textbf{Average}
% \begin{tabular}{lccc}
% \toprule
%         & \textbf{CosSim} & \textbf{L1} & \textbf{RMSE} \\
% \midrule
% \texttt{MXFP4}   & 99.21\% & 0.2350   & 0.4649 \\
% \texttt{NVFP4}   & 99.87\% & 0.0774 & 0.2013 \\
% \bottomrule
% \end{tabular}
% \end{minipage}
% \hfill
% \begin{minipage}{0.48\linewidth}
% \centering
% \textbf{Worst}
% \begin{tabular}{lccc}
% \toprule
%         & \textbf{CosSim} & \textbf{L1} & \textbf{RMSE} \\
% \midrule
% \texttt{MXFP4}   & 98.37\% & 0.2941 & 0.9944 \\
% \texttt{NVFP4}   & 99.52\% & 0.0774 & 0.2013 \\
% \bottomrule
% \end{tabular}
% \end{minipage}
% \label{table:mxfp4}
% \end{table}

\begin{algorithm}[ht]
    \small
    \caption{Implementation of the microscaling \texttt{FP4} attention.}
    \label{alg:sage3_fwd} 
    \begin{algorithmic}[1]
    \STATE {\bfseries Input:} {Matrices $Q(\texttt{FP16}), K(\texttt{FP16}), V(\texttt{FP16}) \in \mathbb{R}^{N \times d}$, block size $B_q, B_{kv}$.}

    \STATE \textbf{Preprocessing:} \jt{$K=K-\mathrm{mean}(K)$} \annotate{// Smoothing K of SageAttention.}
    
    \STATE Divide $Q$ to $T_m = {N}/{B_q}$ blocks $\{\vQ_i\}$; divide $K$, and $V$ to $T_n = {N}/{B_{kv}}$ blocks $\{\vK_i\}$, $\{\vV_i\}$ ;  

    \FOR {$i=1$ {\bfseries to} $T_m$}

      \STATE $\jt{\bar q_i=\mean(\vQ_i),~~ (\mathbf{s_Q}, \hat \vQ_i)} = \jt{\phi(\vQ_i - \bar q_i)}$  ; \annotate{// Smoothing Q of SageAttention2.} 

        \FOR {$j$ in [1, $T_n$]} 
            \STATE $\jt{(\mathbf{s_K}, \hat \vK_j)} = \jt{\phi}(\vK_j^\top)$ , ~~ $\jt{(\mathbf{s_V}, \hat \vV_j)} = \jt{\phi}(\vV_j)$ ; 

            \STATE \jt{$\vS_{ij} = \texttt{FP4MM} (\hat \vQ_i, \mathbf{s_Q}, \hat \vK_j, \mathbf{s_K}) + \texttt{GEMV}(\bar q_i, \vK_j^\top)$} ;  \annotate{// Smoothing Q.}

            \STATE $m_{ij} = \mathrm{max}(m_{i,j-1}, \mathrm{rowmax}(\vS_{ij}))$, $~\widetilde \vP_{ij} = \mathrm{exp}(\vS_{ij} - m_{ij})$, $~l_{ij} = e^{m_{i,j-1}-m_{ij}} l_{i,j-1} + \mathrm{rowsum}(\widetilde \vP_{ij})$ ;

            \STATE \jt{$\mathbf{s_{P_1}} = \rowmax(\widetilde \vP_{ij}) / (448 \times 6)$, ~ $\widetilde \vP_{ij} = \widetilde \vP_{ij} / \mathbf{s_{P_1}}$, ~ $\mathbf{s_{P_2}}, \hat \vP_{ij} = \phi (\widetilde \vP_{ij})$}; \annotate{// two-level quantization}

            \STATE $\vO_{ij} = \mathrm{diag}(e^{m_{i,j-1}-m_{ij}}) \vO_{i,j-1} + \jt{\texttt{FP4MM}(\hat \vP_{ij}, \mathbf{s_{P_2}}, \hat \vV_j, \mathbf{s_V}) \times \mathbf{s_{P_1}}}$
        
        \ENDFOR
        \STATE $\vO_i = \mathrm{diag}(l_{i,T_n}) ^{-1} \vO_{i,T_n}$ ; %~~~ Write $\vO_i$ ;  
    \ENDFOR
    
    \STATE \textbf{return} $O = \{\vO_i\}$
    \end{algorithmic}
    \vspace{-.2em}
\end{algorithm}

\subsection{Two-level Scaling for $\widetilde \vP$}  \label{sec:two_level_quant}
Applying microscaling \texttt{FP4} quantization for $\widetilde \vP$ presents a challenge to attention accuracy. For example, Fig.~\ref{fig:scale_ablation}(~\subref{fig:c}) shows direct quantization severely degrades output quality, producing results substantially different from full-precision outputs. Our analysis reveals that the issue occurs because microscaling \texttt{NVFP4} quantization requires the scale factor to be represented in \texttt{E4M3} \texttt{FP8} format~\citep{micikevicius2022fp8}, rather than the \texttt{FP32} data type typically used for scale factors. This causes accuracy loss when the scale factor is directly converted to \texttt{E4M3} format. 
To better understand this accuracy loss, we analyze the data distribution of $\widetilde \vP$ and its scale factors in Fig.~\ref{fig:quant_errors_of_p}. 
Since $\widetilde \vP$ is computed using online softmax~\citep{milakov2018online}, the values in each microscaling block $\widetilde \vP_{ij}$ fall $\mathsf{[0, 1]}$. Consequently, the scale factor (scale factor = $\max(\widetilde \vP_{ij})/6$) ranges between 0 and 0.167. This narrow range leads to inefficient usage of \texttt{E4M3}'s representable range, increasing accuracy loss. To reduce accuracy loss by fully utilizing \texttt{E4M3}'s range, we propose a two-level quantization method for the $\widetilde \vP$ matrix. Specifically, we first quantize each row of $\widetilde \vP$ to $\mathsf{[0, 448 \times 6]}$. Then we apply the standard \texttt{FP4} quantization $\phi$ for the quantized $\widetilde \vP$. The two-level quantization can be formulated as follows:
\begin{align}
    \mathbf{s_{P_1}} = \rowmax(\widetilde \vP) / (448 \times 6), ~~~
    \widetilde \vP_2 &= \widetilde \vP / \mathbf{s_{P_1}}, ~~~
    \mathbf{s_{P_2}}, \hat \vP_2 = \phi (\widetilde \vP_2) \notag\\
    (\widetilde \vP \approx \hat \vP_2 \times \mathbf{s_{P_2}} \times \mathbf{s_{P_1}}), ~~~~
    \vO &= \texttt{FP4MM}(\hat \vP_2, \mathbf{s_{P_2}}, \hat \vV, \mathbf{s_V}) \times \mathbf{s_{P_1}}
\end{align}
Where $\widetilde \vP$, $\widetilde \vP_2$, and $\mathbf{s_{P_1}}$ are in \texttt{FP32} data type. $\mathbf{s_{P_2}}$ and $\mathbf{s_V}$ are in  \texttt{FP8}  data type. $\hat \vP_2$ and $\hat \vV$ are in \texttt{FP4} data type.

\underline{Empirical results:} As shown in Fig.~\ref{fig:quant_errors_of_p}, our two-level quantization maximizes the \texttt{E4M3} range utilization for $\mathbf{s_P}$, thereby reducing both the numerical representation error of $\mathbf{s_P}$ and the quantization error of $\widetilde{\vP}$. A more formal theoretical analysis is provided in Appendix~\ref{sec:appendix_others}. 
Table~\ref{tab:acc_ablation}(\subref{table:b}) shows the accuracy of two-level quantization against naive direct quantization, using real Q, K, V from layers of CogVideoX. Results indicate that two-level quantization boosts the accuracy.

\subsection{Implementation and Optimization on Hardware}  \label{sec:hardware_implement}

\textbf{Permutation for K.}
Unlike \texttt{FP16}, the \texttt{FP32} accumulator's memory layout in \texttt{FP4} MatMul~\citep{nvidia2024ptx} differs from its operand A's register layout (shown in Fig. \ref{fig:ori_fp32_layout} and \ref{fig:input_layout}). Performing thread shuffles to match operand A's layout would degrade kernel performance.
Our solution transforms the accumulator layout (Fig. \ref{fig:new_fp32_layout}) by permuting the P tile's columns. To maintain correct MatMul, we correspondingly rearrange K's columns, which can be fused with the quantization kernel.

\textbf{Reuse shuffle.} 
The in-kernel micro-scaling quantization of $\widetilde \vP$ requires finding the max value of 16 consecutive row elements. However, as shown in Fig.~\ref{fig:new_fp32_layout}, these 16 elements are distributed across four threads, necessitating intra-thread max reduction followed by inter-thread shuffling, significantly slowing down the kernel.
We optimize this by fusing quantization with online softmax, which also computes row-wise maxima. First, we compute the max over 16 elements in $S$ and reuse it in the subsequent softmax max-reduction. This fusion reduces redundant shuffles and max operations by ~50\%, yielding about 10\% whole kernel speedup.
% The in-kernel micro-scaling quantization of $\widetilde \vP$ requires finding the absolute maximum of 16 consecutive row elements. However, as shown in Fig.~\ref{fig:new_fp32_layout}, these 16 elements are distributed across four threads, necessitating intra-thread absolute max reduction followed by inter-thread shuffling, significantly slowing down the kernel.
% We optimize this by fusing quantization with online softmax, which also computes row-wise maxima and maps all elements in $S$ to be positive. First, we compute the max over 16 elements in $S$. Note that this max need to be exponentiated to be the true value for micro-scaling quantization. Then we resue this maxia for continual online softmax. This fusion reduces redundant shuffles and max operations by ~50\%, with an neligble overhead of one extra exp operation, yielding about 10\% whole kernel speedup.

\textbf{Producer warp epilogue.} 
In conventional warp-specialized kernels, consumer warps typically handle both MatMul and store operations while producers merely load inputs, with ping-pong scheduling between consumers enabling stage overlap~\citep{nvidia2025efficientgemm}. However, register constraints make this approach infeasible for our \texttt{FP4} attention kernel. Instead, we implement ping-pong scheduling between producer warps: while one producer loads inputs for the next MatMul operation, another concurrently stores outputs to global memory, with consumer warps solely responsible for transferring MatMul results from registers to shared memory. This novel design overlaps MatMul and global memory stores within register constraints, boosting throughput.

\section{INT8 Attention for Training}

Low-bit quantization attention works, such as FlashAttention3 and SageAttention, are only for inference. In this section, we propose an \texttt{INT8} attention for training, named \texttt{SageBwd}, which quantizes six of seven matrix multiplications in attention to \texttt{INT8}, achieving no performance degradation in fine-tuning tasks.
Besides, we implement both \texttt{INT8} \texttt{SageBwd} and \texttt{FP8} \texttt{SageBwd} and conduct comparison experiments, proving \texttt{INT8} \texttt{SageBwd} is superior to \texttt{FP8} \texttt{SageBwd} in Section~\ref{sec:in8_vs_fp8_sagebwd}.

\begin{algorithm}[H]
    \small
    \caption{Forward pass of the \texttt{8-bit} attention.}
    \label{alg:int8_train_fwd} 
    \begin{algorithmic}[1]
    \STATE {\bfseries Input:} {\texttt{FP16} matrices $Q, K, V \in \mathbb{R}^{N \times d}$, and block size $B_q, B_{kv}$.}

    \STATE $K_m = {\rm mean}(K); ~~ K\gets K-K_m$ ; \annotate{// Smooth-k technique.}
    
    \STATE Divide $Q$ to $T_m = {N}/{B_q}$ blocks $\{\vQ_i\}$; divide $K$, and $V$ to $T_n = {N}/{B_{kv}}$ blocks $\{\vK_i\}$, $\{\vV_i\}$ ;  

    \STATE {\bfseries Quantization:} \jt{$\{\mathbf{s_Q}, \hat \vQ_i\} = \{\psi (\vQ_i)\}$, ~~ $\{\mathbf{s_K}, \hat \vK_i\} = \{\psi (\vK_i^\top)\}$, ~~$\{\mathbf{s_V}, \hat \vV_i\} = \{\psi (\vV_i)\}$} ; \annotate{// Per-block.}

    \FOR {$i=1$ {\bfseries to} $T_m$}

    \STATE $\vO_i \in \mathbb{R}^{B_q\times D}= (0)$, ~~ $\mathbf{L}_i \in \mathbb{R}^{B_q} = (0),~~ m_i \in \mathbb{R}^{B_{kv}} = (0)$ ;

        \FOR {$j$ in [1, $T_n$]} 

            \STATE \jt{$\vS_{ij} = \texttt{MM} (\hat \vQ_i, \hat \vK_j) \times \mathbf{s_Q} \times \mathbf{s_K}$} ;  

            \STATE $m_{ij} = \mathrm{max}(m_{i,j-1}, \mathrm{rowmax}(\vS_{ij}))$, $\widetilde \vP_{ij} = \mathrm{exp}(\vS_{ij} - m_{ij})$, $l_{ij} = e^{m_{i,j-1}-m_{ij}} l_{i,j-1} + \mathrm{rowsum}(\widetilde \vP_{ij})$;

            \STATE \jt{$\mathbf{s_P} = \mathrm{exp}(\rowmax(\vS_{ij}) - m_{ij}) / 127$, ~~ $\mathbf{\hat \vP}_{ij} = \widetilde \vP_{ij} / \mathbf{s_P}$} ; \annotate{// Per-token quantization.}

            \STATE $\vO_{ij} = \mathrm{diag}(e^{m_{i,j-1}-m_{ij}}) \vO_{i,j-1} + \jt{\texttt{MM}(\hat \vP_{ij}, \hat \vV_j) \times \mathbf{s_{P}} \times \mathbf{s_V}}$
        
        \ENDFOR
        \STATE $\vO_i = \mathrm{diag}(l_{i,T_n}) ^{-1} \vO_{i,T_n}$ ;  %~~~ Write $\vO_i$ ;  
        \STATE $\mathbf{L}_i = m_{i, T_n} + \mathrm{log}(l_{i, T_n})$ ; %~~~ Write $\mathbf{L}_i$ ;
    \ENDFOR
    
    \STATE \textbf{return} $O = \{\vO_i\}$, $L = \{\mathbf{L}_i\}$ ;
    \end{algorithmic}
    \vspace{-.15em}
\end{algorithm}

\subsection{Forward}
There are two matrix multiplications in the forward pass of attention:
\begin{align}
    \vS = \vQ\vK^\top,~~ \vO = \vP \vV
\end{align}
\textbf{Per-token quantization for P.} Following SageAttention~\citep{2024sageattention}, we apply smoothing K and per-block \texttt{INT8} quantization for the $\vQ\vK^\top$. However, for the $\widetilde \vP \vV$, a static per-block \texttt{INT8} quantization with a static scale factor of $\mathsf{1/127}$ for $\widetilde \vP$ is inaccurate~\citep{2024sageattention}. Fortunately, we find applying per-token \texttt{INT8} quantization for $\widetilde \vP \vV$ and per-block \texttt{INT8} quantization for $\vV$ can enhance the attention accuracy. 
Furthermore, we eliminate the need for explicit $\max$ operations on $\vP$ by reusing both global and local maximum values from the online softmax computation (Line 9 in Algorithm~\ref{alg:int8_train_fwd}).
The algorithm for the forward is shown in Algorithm~\ref{alg:int8_train_fwd}.

Given our extensive use of \texttt{INT8} per-block quantization in trainable attention, we formalize the process as follows. For each FlashAttention block $\mathbf{X}$, the quantization process $\mathbf{s_X}, \mathbf{\hat X} = \psi(\mathbf{X})$ can be formulated as:
\begin{align}
    \mathbf{s_X} = \max(|\mathbf{X}|)/127, ~~~ \mathbf{\hat X}=\mathbf{X}/\mathbf{s_X}
\end{align}
\begin{algorithm}[H]
    \small
    \caption{Backward pass of the \texttt{8-bit} attention.}
    \label{alg:int8_train_bwd} 
    \begin{algorithmic}[1]
    \STATE {\bfseries Input:} \jt{$\{\mathbf{s_Q}, \hat \vQ_i\}, \{\mathbf{s_K}, \hat \vK_i\}, \{\mathbf{s_V}, \hat \vV_i\}$}, $K_m$, $O$, $\{\mathbf{L}_i\}$ from forward, $dO \in \mathbb{R}^{N \times d}$, block size $B_q, B_{kv}$ ; 

    \STATE $D = \mathrm{rowsum}(dO \circ O)$, ~~ divide $D$ to $T_m = {N}/{B_q}$ blocks $\{\mathbf{D}_i\}$; 

    \FOR {$j=1$ {\bfseries to} $T_n$}

        \FOR {$i$ in [1, $T_m$]} 

            \STATE \jt{$\vS_{ij} = \texttt{MM} (\hat \vQ_i, \hat \vK_j) \times \mathbf{s_Q} \times \mathbf{s_K}$} ;  ~~~~~$\vP_{ij} = \mathrm{exp}(\vS_{ij} - \mathbf{L}_{i})$ ;

            \STATE \jt{$\mathbf{s_P}, \hat \vP_{ij} = \psi(\vP_{ij})$, ~~ $\mathbf{s_{dO}}, \hat \vdO_{i} = \psi(\vdO_{i})$} ; ~\annotate{// \texttt{INT8} per-block quantization.}

            \STATE $\vdV_j \leftarrow \vdV_j + \jt{\texttt{MM}(\hat \vP_{ij}^\top, \hat \vdO_{i}) \times \mathbf{s_P} \times \mathbf{s_{dO}}}$ ;

            \STATE $\vdP_{ij} = \texttt{MM}(\vdO, \vV_j^\top)$ ;  ~\annotate{// Keep in \texttt{FP16}.}

            \STATE $\vdS_{ij} = \vP_{ij} \circ (\vdP_{ij} - \mathbf{D}_i)$ ; ~~~~~\jt{$\mathbf{s_{dS}}, \hat \vdS_{ij} = \psi(\vdS_{ij})$} ; ~\annotate{// \texttt{INT8} per-block quantization.}

            \STATE $\vdQ_i \leftarrow \vdQ_i + \jt{\texttt{MM}(\hat \vdS_{ij}, \hat \vK_{j}) \times \mathbf{s_{dS}} \times \mathbf{s_K}} + {\rm rowsum}(\vdS_{ij}) K_m$ ; \annotate{// Backward for smooth-k.}

            \STATE $\vdK_j \leftarrow \vdK_j + \jt{\texttt{MM}(\hat {\vdS}^\top_{ij}, \hat \vQ_{i}) \times \mathbf{s_{dS}} \times \mathbf{s_Q}}$ ;

        \ENDFOR

        % Write $\vdK_j$ and $\vdV_j$ ;  
    \ENDFOR
    
    \STATE \textbf{return} $dQ, dK, dV$ ;
    \end{algorithmic}
    \vspace{-.15em}
\end{algorithm}
\subsection{Backward}

There are five matrix multiplications in the backward pass of attention:
\begin{align}
    \vS = \vQ\vK^\top,~~~
    % \widetilde \vP = \texttt{OnlineSoftmax}(\vS) \\
    \vdV = \widetilde \vP^\top \vdO,~~~
    \vdP = \vdO \vV^\top,~~~
    \vdQ = \vdS \vK,~~~
    \vdK = \vdS^\top \vQ
\end{align}
We observe that whether applying quantizing to $\vdO \vV^\top$ has a significant impact on the accuracy of the gradient of $Q, K$. This is because the accuracy of $\vdO \vV^\top$ directly determines the accuracy of $\vdP$ and $\vdS$ (see computational dependencies in Algorithm~\ref{alg:int8_train_bwd}). The accuracy loss in $\vdS$ will continuously accumulate errors into $\vdQ$ and $\vdK$ during the recurrent process along the sequence length in FlashAttention's backward pass, meaning longer sequences lead to greater error accumulation. 
Therefore, we maintain $\vdO \vV^\top$ in \texttt{FP16} while accelerating the other four matrix multiplications using \texttt{INT8} per-block quantization.  The algorithm for the forward is shown in Algorithm~\ref{alg:int8_train_bwd}. 
\underline{Empirical results:} Table~\ref{tab:acc_ablation} (c) shows the accuracy of the $\vdQ$ with and without quantization of $\vdO \vV^\top$. We find that the accuracy of $\vdQ$ is significantly improved when keeping $\vdO \vV^\top$ in \texttt{FP16}.

\begin{table}[H]
\vspace{-.5em}
\footnotesize
\centering
% \sisetup{detect-weight=true, detect-family=true}
\caption{Accuracy ablation using different quantization strategies.} % 主标题
\label{tab:acc_ablation}
\begin{minipage}{\linewidth} % 保证三个子表格同宽
\centering
\begin{tabular}{@{}ccc@{}}
% 子表格 (a)
\begin{minipage}{0.3\linewidth}
\centering
\setlength{\tabcolsep}{3pt}
\subcaption{Different \texttt{FP4} choices} % 子标题
\label{table:a}
\begin{tabular}{@{}cccc@{}}
\toprule
\textbf{Type} & \textbf{CosSim}$\uparrow$ & \textbf{L1}$\downarrow$ & \textbf{RMSE}$\downarrow$ \\
\midrule
\texttt{MXFP4} & 98.37\% & 0.294 & 0.994 \\
\texttt{NVFP4} & \textbf{99.52\%} & \textbf{0.077} & \textbf{0.201} \\
\bottomrule
\end{tabular}
\end{minipage}
&
% 子表格 (b)
\begin{minipage}{0.31\linewidth}
\centering
\setlength{\tabcolsep}{3pt}
\subcaption{Different scale strategies for $\widetilde \vP$} 
\label{table:b}
\begin{tabular}{@{}cccc@{}}
\toprule
\textbf{Method} & \textbf{CosSim} & \textbf{L1} & \textbf{RMSE} \\
\midrule
Direct          & 93.32\% & 0.193 & 1.103 \\
Two-level   & \textbf{99.52\%} & \textbf{0.077} & \textbf{0.201} \\
\bottomrule
\end{tabular}
\end{minipage}
&
% 子表格 (c)
\begin{minipage}{0.31\linewidth}
\centering
\setlength{\tabcolsep}{3pt}
\subcaption{Different data types for $\vdO \vV^\top$}
\label{table:c}
\begin{tabular}{@{}cccc@{}}
\toprule
\textbf{Method} & \textbf{CosSim} & \textbf{L1} & \textbf{RMSE} \\
\midrule
\texttt{INT8}  & 97.47\% & 0.171 & 2.440 \\
\texttt{FP16}  & \textbf{99.77\%} & \textbf{0.039} & \textbf{0.692} \\
\bottomrule
\end{tabular}
\end{minipage}
\end{tabular}
\end{minipage}
\vspace{-.5em}
\end{table}

\begin{figure}[H]
    \centering
    \includegraphics[width=\columnwidth]{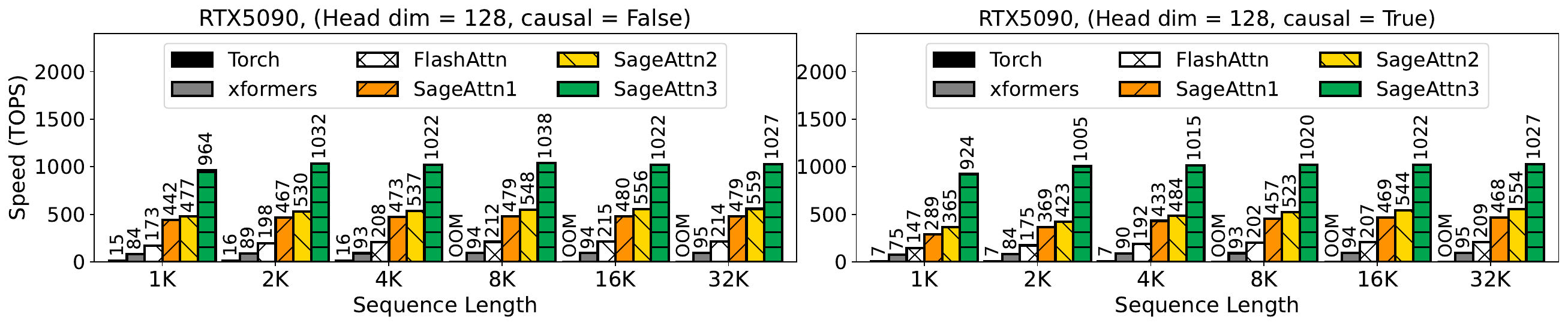} 
    \vspace{-1.5em}
    \caption{Speed comparison between \our and Baselines (\texttt{RTX5090}, headim=128).}
    \vspace{-1em}
    \label{fig:sage3_h128_speed} 
\end{figure}

\begin{figure}[H]
    \centering
    \includegraphics[width=\columnwidth]{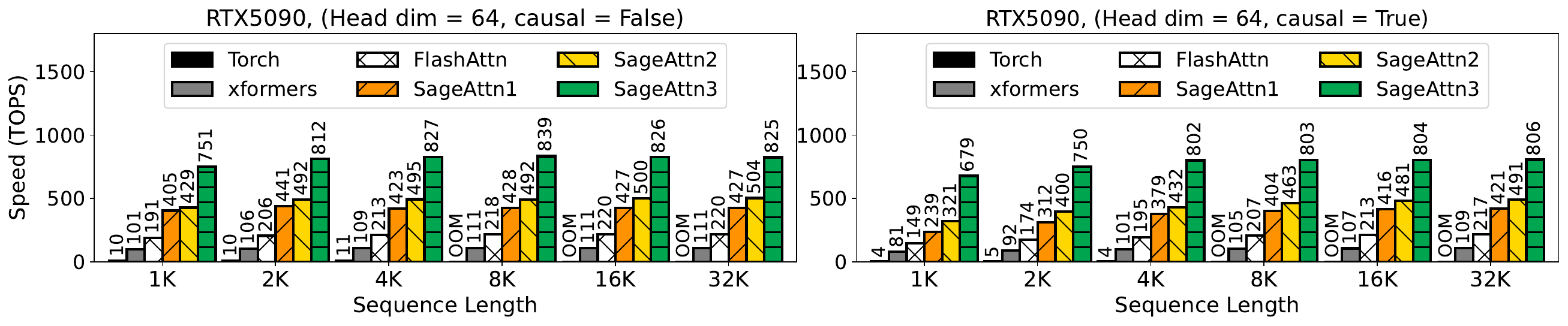} 
    \vspace{-1.5em}
    \caption{Speed comparison between \our and Baselines (\texttt{RTX5090}, headim=64).}
    \vspace{-1em}
    \label{fig:sage3_h64_speed} 
\end{figure}

\begin{figure}[H]
  \centering
  \includegraphics[width=\columnwidth]{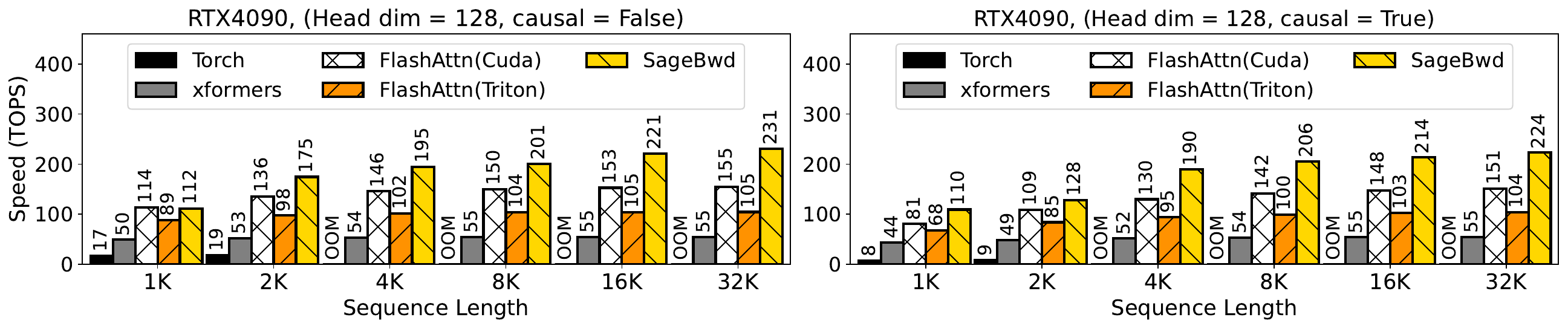} 
  \vspace{-1.5em}
  \caption{Speed comparison between \texttt{SageBwd} and Baselines (\texttt{RTX4090}, headim=128).}
  \vspace{-1em}
  \label{fig:sage_train_h128_speed} 
\end{figure}

\begin{figure}[H]
    \centering
    \includegraphics[width=\columnwidth]{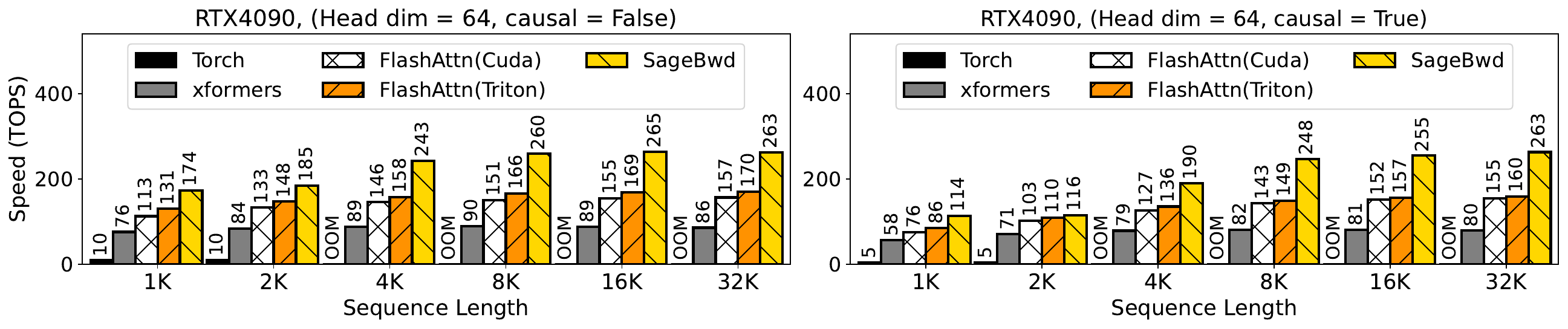}
    \vspace{-1.5em}
    \caption{Speed comparison between \texttt{SageBwd} and Baselines (\texttt{RTX4090}, headim=64).}
    \vspace{-1em}
    \label{fig:sage_train_h64_speed} 
\end{figure}

\begin{table}[H]  
\vspace{-1em}
        \caption{End-to-end metrics comparison on various models.}
        \sisetup{detect-weight=true, detect-family=true}
        \sisetup{table-auto-round=true}
        \label{tab:end2end_metric}
        \begin{center}
        \setlength\tabcolsep{7pt}
        \scalebox{0.89}{\begin{tabular}{p{1.3cm}
                                        p{3.5cm}
                                        c
                                        c
                                        c
                                        c
                                        c}
        \toprule
        {\bf Model}  & {\mbox{\hspace{2em}\textbf{Attention}}}  & {\bf CLIPSIM $\uparrow$}  & {\bf CLIP-T $\uparrow$}  & {\bf VQA-a $\uparrow$}  & {\bf VQA-t $\uparrow$}  & {\bf FScore $\uparrow$} \\ \hline
    
        \multirow{3}{*}{\mbox{\hspace{-.75em}\makecell[c]{\cogvideo}}} 
& \mbox{\hspace{-.63em}Full-Precision \texttt{(16bit)}}  & 0.1865 & 0.9968 & 70.476 & 69.875 & 4.780 \\
% & \mbox{\hspace{-.63em}\texttt{SageAttention2 (6bit)}}  &  &  &  &  &  \\
& \mbox{\hspace{-.63em}\texttt{SageAttention2 (8bit)}}  & 0.1880 & 0.9969 & 69.414 & 70.750 & 4.534 \\
& \mbox{\hspace{-.63em}\ourf \texttt{(4bit)}}  & \textbf{0.1881} & \textbf{0.9969} & \textbf{69.860} & \textbf{70.364} & \textbf{4.035} \\

         \hline

        \multirow{3}{*}{\mbox{\hspace{-.3em}\makecell[c]{\texttt{Hunyuan}\\ \texttt{Video}}}} 
        & \mbox{\hspace{-.63em}Full-Precision \texttt{(16bit)}}  & 0.1838 & 0.9993 & 68.998 & 78.891 & 1.4793 \\ 
        % & \mbox{\hspace{-.63em}\texttt{SageAttention2 (6bit)}}  &  &  &  &  &  \\
        & \mbox{\hspace{-.63em}\texttt{SageAttention2 (8bit)}}  & 0.1836 & 0.9993 & 69.497 & 77.019 & 1.4741 \\
        & \mbox{\hspace{-.63em}\ourf \texttt{(4bit)}}  & \textbf{0.1866} & \textbf{0.9993} & \textbf{70.552} & \textbf{75.440} & \textbf{1.232} \\ 
        \hline

        \multirow{3}{*}{\mbox{\hspace{.1em}\makecell[c]{\mochi}}} 
        & \mbox{\hspace{-.63em}Full-Precision \texttt{(16bit)}}  & 0.1828 & 0.9990 & 61.9840 & 61.0000 & 1.8042 \\ 
        % & \mbox{\hspace{-.63em}\texttt{SageAttention2 (6bit)}}  &  &  &  &  &  \\
        & \mbox{\hspace{-.63em}\texttt{SageAttention2 (8bit)}}  & 0.1819 & 0.9990 & 61.0093 & 60.3732 & 1.7539 \\
        & \mbox{\hspace{-.63em}\ourf \texttt{(4bit)}}  & \textbf{0.1800} & \textbf{0.9993} & \textbf{61.863} & \textbf{59.429} & \textbf{1.649} \\ 
        \bottomrule

        \end{tabular} }
        \end{center}
    
    \vspace{-.75em}
    
        \begin{center}
        \setlength\tabcolsep{19.45pt}
        \scalebox{0.89}{\begin{tabular}{p{0.51cm}
                                        p{2.6cm}
                                        c
                                        c
                                        c
                                        c}
        \toprule
        {\mbox{\hspace{-.75em}\textbf{Model}}}  & {\bf Attention}  & {\bf FID $\downarrow$}  & {\bf sFID $\downarrow$}  & {\bf CLIP $\uparrow$}  & {\bf IR $\uparrow$}
        \\ \hline

        \multirow{3}{*}{\hspace{-1em}\flux} 
        & \mbox{\hspace{-1.8em}Full-Precision \texttt{(16bit)}} & 162.812 & 146.980 & 31.409 & 0.91 \\ 
        & \mbox{\hspace{-1.8em}\texttt{SageAttention2 (8bit)}} & 163.107 & 146.213 & 31.436 & 0.90 \\
        & \mbox{\hspace{-1.8em}\our \texttt{(4bit)}} & \textbf{162.121} & \textbf{142.839} & \textbf{31.450} & \textbf{0.94} \\
        \hline

        \multirow{3}{*}{\hspace{-2.1em}\texttt{\makecell[c]{Stable-Di\\ffusion3.5}}} 
        & \mbox{\hspace{-1.8em}Full-Precision \texttt{(16bit)}} & 166.421 & 146.379 & 31.93 & 0.93 \\ 
        & \mbox{\hspace{-1.8em}\texttt{SageAttention2 (8bit)}} & 164.986 & 148.557 & 32.01 & 0.93 \\
        & \mbox{\hspace{-1.8em}\ourf \texttt{(4bit)}} & \textbf{166.102} & \textbf{145.587} & \textbf{32.01} & \textbf{0.92} \\
 
        \bottomrule
    \end{tabular} }
    \end{center}
    \vspace{-.5em}
\end{table}

\begin{figure}[H]
\vspace{-1em}
  \centering
  \begin{minipage}[b]{0.19\textwidth}
    \includegraphics[width=\linewidth]{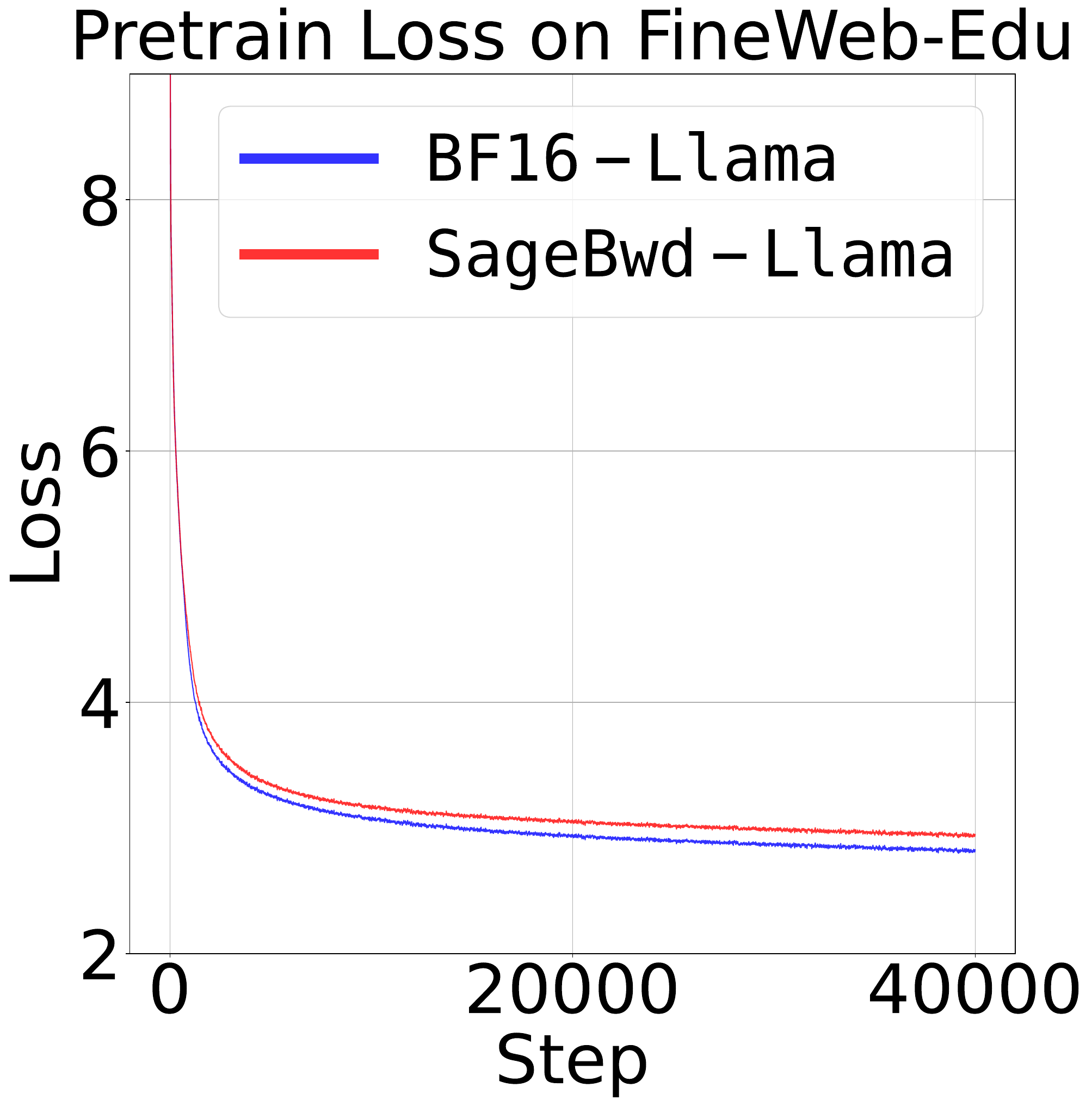}
    \subcaption{}
    \label{fig:train_a}
  \end{minipage}
  \begin{minipage}[b]{0.187\textwidth}
    \includegraphics[width=\linewidth]{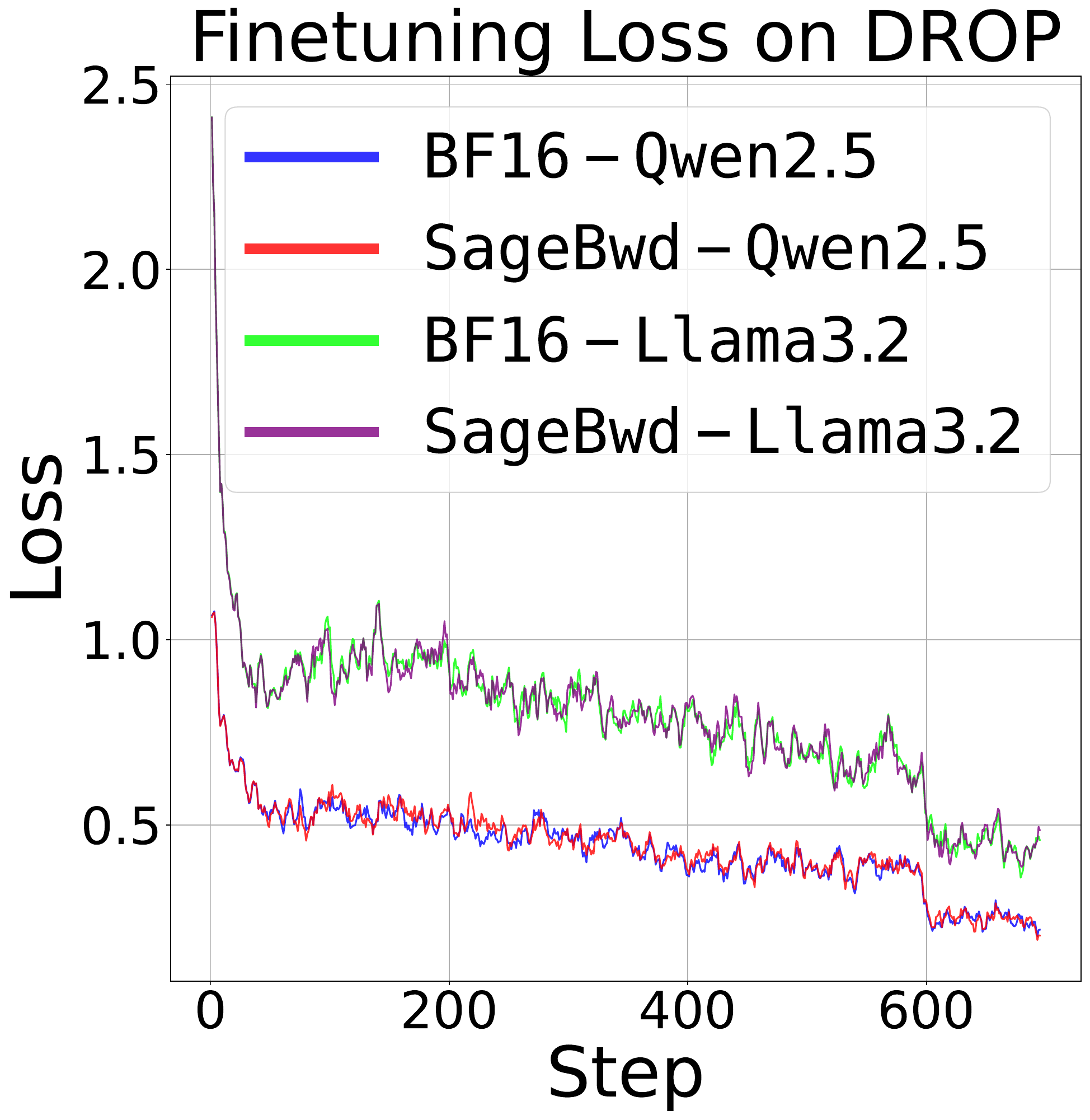}
    \subcaption{}
    \label{fig:train_b}
  \end{minipage}
  \begin{minipage}[b]{0.193\textwidth}
    \includegraphics[width=\linewidth]{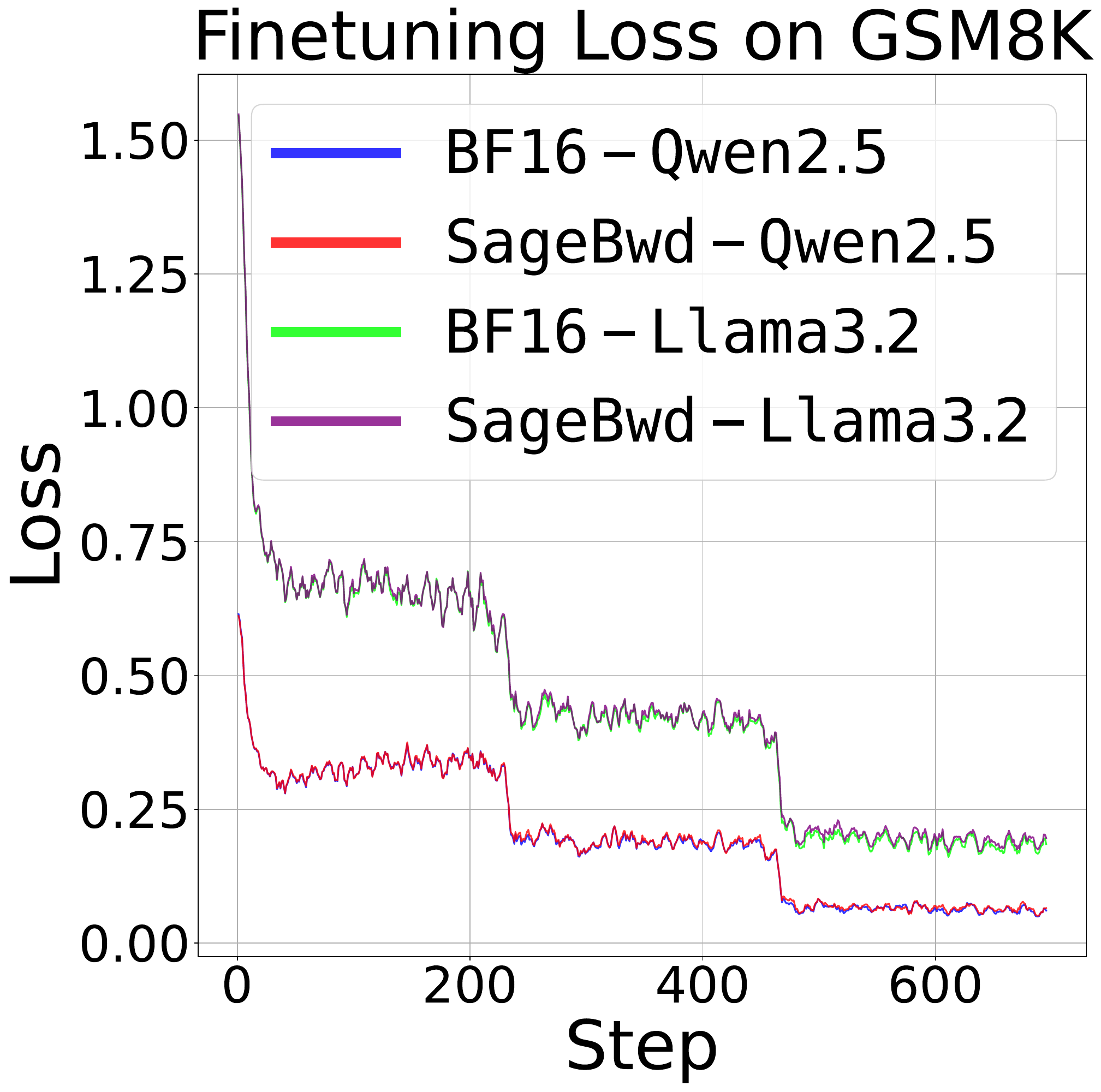}
    \subcaption{}
    \label{fig:train_c}
  \end{minipage}
  \begin{minipage}[b]{0.203\textwidth}
    \includegraphics[width=\linewidth]{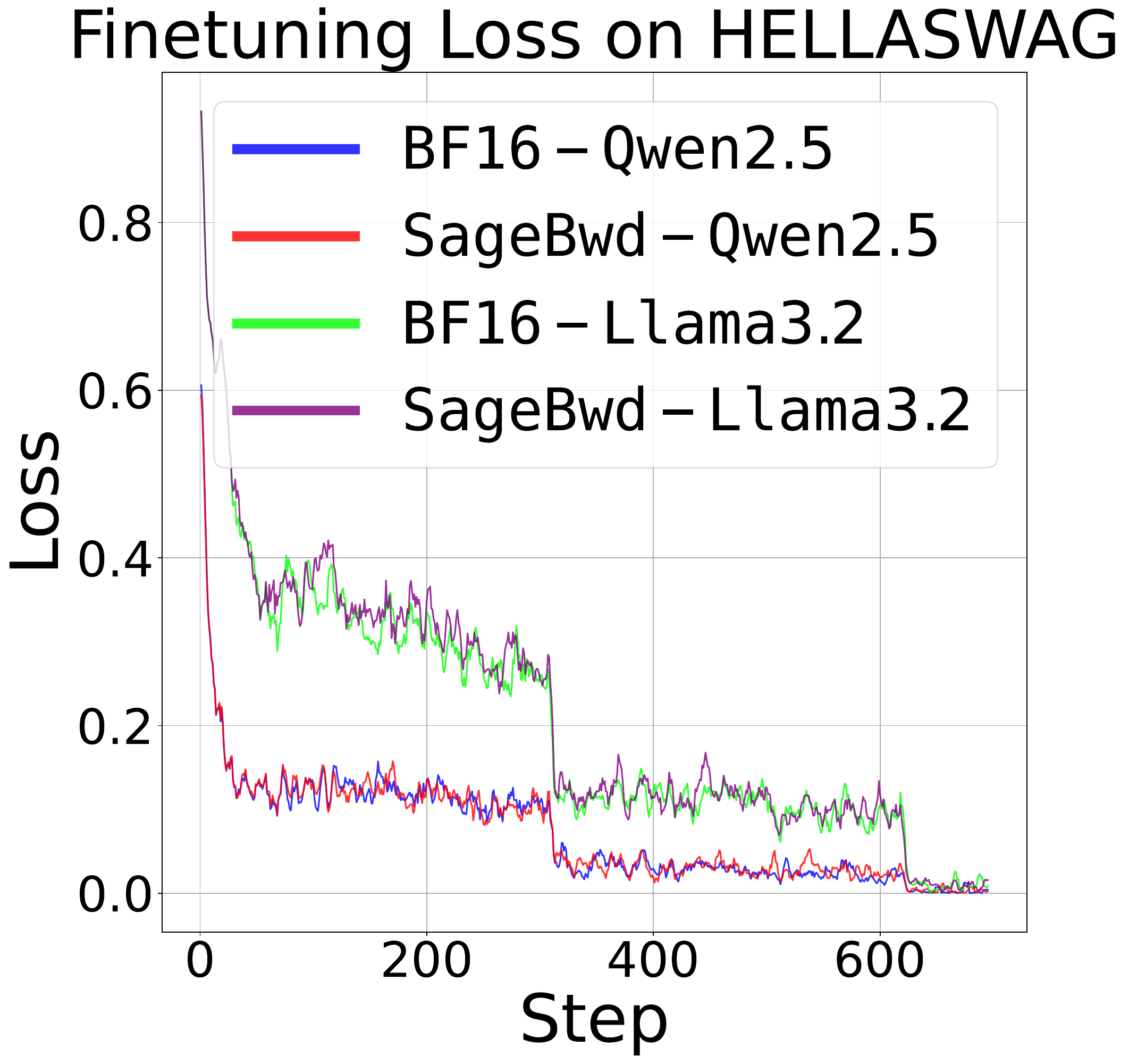}
    \subcaption{}
    \label{fig:train_d}
  \end{minipage}
  \begin{minipage}[b]{0.187\textwidth}
    \includegraphics[width=\linewidth]{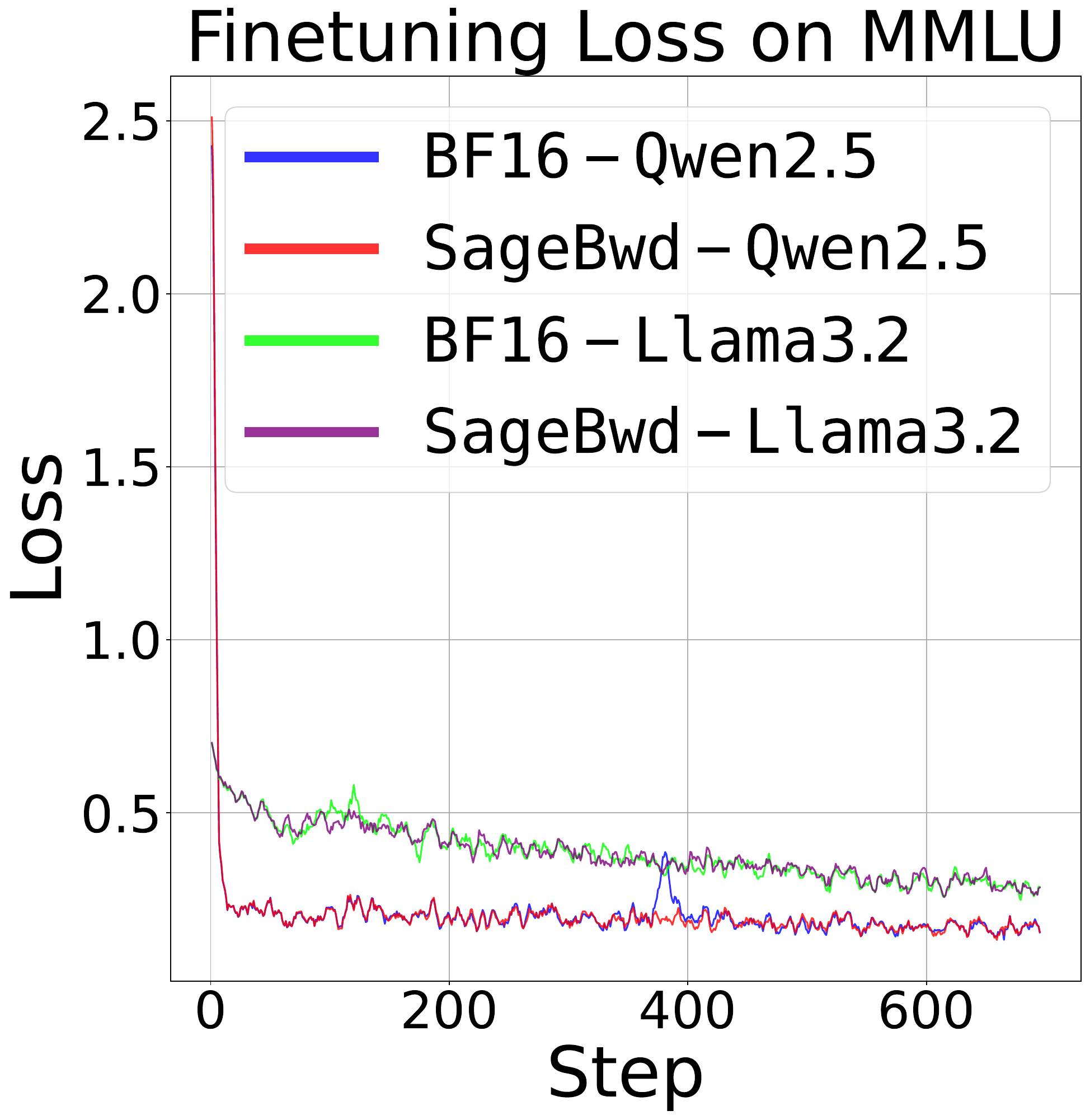}
    \subcaption{}
    \label{fig:train_e}
  \end{minipage}
  \vspace{-.45em}
  \caption{Pretraining and Finetuing loss curves of \texttt{BF16} and \texttt{8-bit} atttention.}
  \vspace{-.5em}
  \label{fig:finetune-loss}
\end{figure}

\begin{table}[H] 
% \small
\centering
\caption{\texttt{8-bit} attention finetune results on \qwen and \llamal models.}
\label{tab:finetune}
% \scriptsize
\scalebox{0.96}{
\begin{tabular}{
    cc 
    cc
    cc
    cc
    cc
    cc
}
\toprule
{Model} 
& {Method} 
& {GSM8K}\tiny{(Acc$\uparrow$)} 
& {DROP}\tiny{(F1$\uparrow$)}
& {MMLU}\tiny{(Acc$\uparrow$)} 
& {HELLASWAG}\tiny{(Acc$\uparrow$)} \\
\midrule
\multirow{2}{*}{\qwen(1.5B)} & \texttt{BF16} 
& 0.521 & 0.733 & 0.569 & 0.905 \\
\cmidrule{2-6}
& \texttt{SageBwd} 
& 0.520 & \textbf{0.734} & \textbf{0.574} & \textbf{0.911} \\

\midrule
\multirow{2}{*}{\qwen(3B)} & \texttt{BF16} 
& 0.601 & 0.785 & 0.640 & 0.944 \\
\cmidrule{2-6}
& \texttt{SageBwd}  
& \textbf{0.607} & 0.782 & \textbf{0.653} & 0.943 \\

\midrule
\multirow{2}{*}{\llamal(1B)} & \texttt{BF16} 
& 0.259 & 0.641 & 0.464 & 0.828 \\
\cmidrule{2-6}
& \texttt{SageBwd}   
& \textbf{0.268} & 0.637 & 0.458 & 0.823 \\
\bottomrule
\end{tabular}}
% \vspace{-1em}
\end{table}

\section{Experiments}
\textbf{Main results.} \our is faster than FlashAttention and xformers by 5$\times$ and 11$\times$ on \texttt{RTX5090}, and maintains end-to-end metrics across various models. Furthermore, \texttt{SageBwd} is faster than FlashAttention and xformers by 1.67$\times$ and 3$\times$ on \texttt{RTX4090}, and achieves no measurable degradation in fine-tuning tasks.

\subsection{Setup}
\noindent \noindent\textbf{Models and attentions.} 
We validate the effectiveness of \our and \sageback across a diverse set of representative models from language, image, and video generation.
Specifically, we conduct experiments on: \qwen~\citep{yang2024qwen2} and \llamal~\citep{llama31model} for text2text, \cogvideo~\citep{yang2024cogvideox}, \hyvideo~\citep{kong2024hunyuanvideo}, and \mochi~\citep{genmo2024mochi} for text2video, \flux~\citep{flux}, and \sd~\citep{stable_diffusion_3_5} for text2image. 
We compare our method with FlashAttention2~\citep{dao2023flashattention}, xformers~\citep{xFormers2022}, SageAttention~\citep{2024sageattention}, and SageAttention2~\citep{zhang2024sageattention2}. Please \textbf{note} that FlashAttention3 can only run on Hopper GPUs, so FlashAttention2 is already the fastest version for \texttt{RTX5090} and \texttt{RTX4090}. 

\noindent \noindent\textbf{Datasets, metrics, and hyperparameters.} For the details about the datasets, metrics, and hyperparameters we used, please refer to Appendix~\ref{sec:exp_dataset_metrics}.

\noindent\textbf{Implementation.} We implement \our using CUTLASS~\citep{cutlass} and CUDA, and implement \sageback using OpenAI Triton~\citep{openaitriton}.

\begin{figure}[H]
  \centering
  \includegraphics[width=\columnwidth]{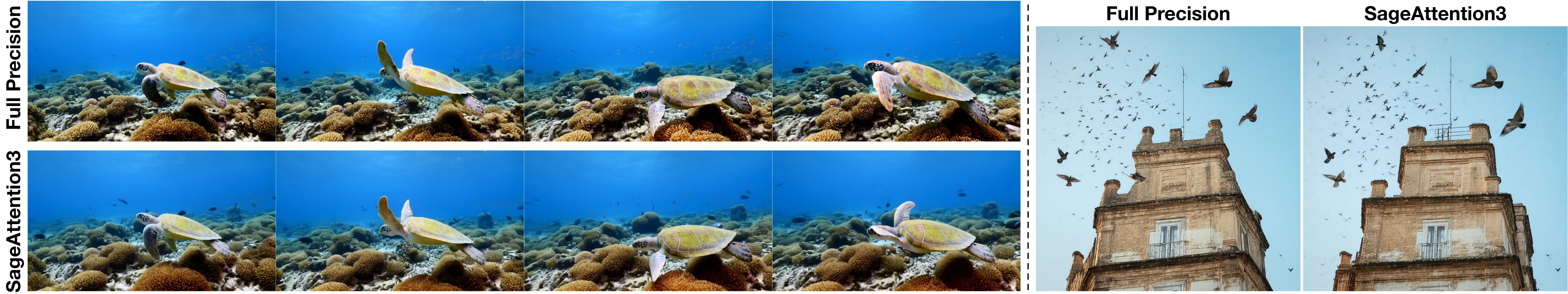} 
  \vspace{-1em}
  \caption{Visible examples of video generation on \texttt{HunyuanVideo} (left) and image generation on \texttt{Stable-Diffusion3.5} (right).}
  \vspace{-.5em}
  \label{fig:visual_example} 
\end{figure}

\begin{table}[H]
\sisetup{detect-weight=true, detect-family=true}
\centering
\vspace{-.5em}
\caption{End-to-end speedup performance using \our and \sageback.}
\label{tab:e2e_latency}
\begin{subtable}[t]{0.5\textwidth}
    \centering
    \subcaption{Inference latency using \our.}
    \label{tab:e2e_inference_latency} % Label should be after subcaption
    \scalebox{0.95}{
    \begin{tabular}{@{}ccccc@{}}
    \toprule
    \textbf{Model} &  \textbf{Original} & \texttt{Sage1} & \texttt{Sage2} & \texttt{Sage3}  \\
    \midrule
    \texttt{CogvideoX} (2B)   & \num{64}~s    &  \num{55}~s  & \num{46}~s & \textbf{\num{27}~s}  \\
    \texttt{HunyuanVideo}       & \num{489}~s  & \num{257}~s & \num{240}~s & \textbf{\num{164}~s} \\
    \bottomrule
    \end{tabular}}
\end{subtable}
\hfill
\begin{subtable}[t]{0.45\textwidth}
    \centering
    \subcaption{One iteration training latency using \texttt{SageBwd}.}
    \label{tab:e2e_training_latency} % Label should be after subcaption
    \scalebox{0.95}{
    \begin{tabular}{@{}ccc@{}}
    \toprule
    \textbf{Model}  & \textbf{Original} &  \texttt{SageBwd} \\
    \midrule
    \texttt{Llama} (8K)  & \num{2.1}~s  &\textbf{\num{1.9}~s} \\
    \texttt{Llama} (16K)  & \num{6.0}~s   &\textbf{\num{5.2}~s} \\
    \bottomrule
    \end{tabular}}
\end{subtable}
\end{table}

\subsection{Efficiency and Effectiveness}
\noindent\textbf{Kernel Speed.} Fig.~\ref{fig:sage3_h128_speed} and~\ref{fig:sage3_h64_speed} show the kernel speed of \our and baselines on \texttt{RTX5090}. We can see that \our achieves 4\textasciitilde \textbf{5}$\times$ speedup over FlashAttention2 and 8\textasciitilde \textbf{11}$\times$ speedup over xformers. Fig.~\ref{fig:sage_train_h128_speed} and~\ref{fig:sage_train_h64_speed} show the forward+backward speed of \sageback and baselines on \texttt{RTX4090}. It shows that \sageback achieves \textbf{1.67}$\times$ speedup at most than FlashAttention2 and a higher speedup than FlashAttention2 implemented in Triton and xformers.

\noindent\textbf{End-to-end metrics loss of SageAttention3.} In Table~\ref{tab:end2end_metric}, we compare the end-to-end quality metrics on various models using \our and other attention methods. The results demonstrate that \our almost incurs almost no end-to-end quality loss across these models.

\noindent\textbf{End-to-end metrics loss of SageBwd.} To evaluate the effectiveness of \sageback on training tasks, we conduct two experiments. First, we fine-tune the base models of \qwen(3B) and \llamal(1B) on GSM8K~\citep{cobbe2021gsm8k}, DROP~\citep{Dua2019DROP}, MMLU~\citep{MMLU}, and HELLASWAG~\citep{zellers2019hellaswag} datasets. Fig.~\ref{fig:finetune-loss} (\subref{fig:train_b}-\subref{fig:train_e}) shows the fine-tuning loss results, indicating that \sageback perfectly aligns with \texttt{BF16}. Moreover, our evaluation of the fine-tuned models' answer quality across multiple test datasets (Table~\ref{tab:finetune}) demonstrates that \sageback achieves the same performance as \texttt{BF16}. 
Second, we conduct pre-training tasks on FineWeb-Edu~\citep{lozhkov2024fineweb-edu} using a \texttt{Llama} (400M)~\citep{touvron2023llama} model. Fig.~\ref{fig:finetune-loss} (\subref{fig:train_a}) shows the loss curve, indicating that while \sageback can achieve loss convergence, its convergence speed is relatively slow. This limitation restricts its applicability in pretraining tasks.

\noindent\textbf{Visible example.} Fig.~\ref{fig:visual_example} visualizes some comparative examples of video generation on \texttt{HunyuanVideo} and image generation on \texttt{Stable-diffsion3.5} using \our. The results demonstrate that \our maintains full generation quality. Additional visible examples are provided in Fig.~\ref{fig:sd_visual},~\ref{fig:flux_visual},~\ref{fig:cog_visual}, and~\ref{fig:hunyuan_visual} in the Appendix.

\noindent\textbf{End-to-end speedup.}
Table~\ref{tab:e2e_latency}(\subref{tab:e2e_inference_latency}) and~\ref{tab:e2e_latency}(\subref{tab:e2e_training_latency}) summarize end-to-end inference and training latency improvements. The results show that~\our (Table~\ref{tab:e2e_latency}(\subref{tab:e2e_inference_latency})) achieved about \textbf{3}$\times$ (\texttt{HunyuanVideo}) and \textbf{2.4}$\times$ (\texttt{CogVideoX}) end-to-end inference generation speedups on \texttt{RTX5090}. Furthermore, \sageback (Table~\ref{tab:e2e_latency}(\subref{tab:e2e_training_latency})) accelerates the training of Llama (1B) by about \textbf{1.15}$\times$ using 8K/16K token micro-batches on \texttt{RTX4090}.

\subsection{Benefit of Using Both SageAttention3 and SageBwd}
\vspace{-5mm}
\begin{table}[H]
\sisetup{detect-weight=true, detect-family=true}
\centering
\vspace{-.7em}
\caption{Comparison between \texttt{BF16} and \texttt{INT8} fine-tuning followed by \texttt{FP4} inference.}
\label{tab:int8_fp4_combined}
\begin{subtable}[t]{0.48\textwidth}
    \centering
    \subcaption{\texttt{Qwen2.5-1.5B} results.}
    \label{tab:qwen15b_fp4}
    \scalebox{0.95}{
    \begin{tabular}{@{}lcc@{}}
    \toprule
    \textbf{Method} & \textbf{GSM8k} $\uparrow$ & \textbf{MMLU} $\uparrow$ \\
    \midrule
    \texttt{BF16} Fine-tuning & 0.4912 & 0.4688 \\
    \textbf{\texttt{SageBwd} Fine-tuning} & \textbf{0.5232} & \textbf{0.4934} \\
    \bottomrule
    \end{tabular}}
\end{subtable}
\hfill
\begin{subtable}[t]{0.48\textwidth}
    \centering
    \subcaption{\texttt{Qwen2.5-3B} results.}
    \label{tab:qwen3b_fp4}
    \scalebox{0.95}{
    \begin{tabular}{@{}lcc@{}}
    \toprule
    \textbf{Method} & \textbf{GSM8k} $\uparrow$ & \textbf{MMLU} $\uparrow$ \\
    \midrule
    \texttt{BF16} Fine-tuning & 0.5860 & 0.6000 \\
    \textbf{\texttt{SageBwd} Fine-tuning} & \textbf{0.5945} & \textbf{0.6032} \\
    \bottomrule
    \end{tabular}}
\end{subtable}
\vspace{-0.7em}
\end{table}

We first apply \texttt{SageBwd} during fine-tuning, followed by \texttt{SageAttention3} during inference. Specifically, we fine-tuned \texttt{Qwen2.5} for 1,000 steps using either \texttt{BF16} or \texttt{SageBwd}, and then evaluated inference performance using \texttt{SageAttention3}. The results on GSM8k and MMLU are shown in Table~\ref{tab:int8_fp4_combined}, \texttt{INT8} \texttt{SageBwd} fine-tuning followed by \texttt{FP4} \texttt{SageAttention3} inference achieves higher accuracy on GSM8k and MMLU, suggesting the approaches are complementary. This improvement is likely because \texttt{INT8} and \texttt{FP4} share a more similar representable data distribution, reducing the mismatch error compared to \texttt{BF16}.

\subsection{INT8 SageBwd vs FP8 SageBwd}
\label{sec:in8_vs_fp8_sagebwd}
% \vspace{-2.5em}
We choose \texttt{INT8} for \texttt{SageBwd} for two key reasons: (1) Higher gradient accuracy in attention backward.
The backward of \texttt{INT8} attention yields more accurate gradients for $Q$, $K$, and $V$ compared to \texttt{FP8}.
We evaluate all layers of \texttt{CogVideoX-2B} and report the L1 error and cosine similarity of the gradients in Table~\ref{tab:int8_vs_fp8_l1} and Table~\ref{tab:int8_vs_fp8_cos}.
For fairness, $\mathbf{dOV}^\top$ is kept in \texttt{FP16} for both methods.
As shown in the results, \texttt{INT8 SageBwd} achieves lower L1 error and higher cosine similarity than \texttt{FP8 SageBwd}. (2) Wider hardware support.
\texttt{INT8} is supported on almost all modern GPUs, including NVIDIA \texttt{A100} and many non-NVIDIA devices (e.g., AMD \texttt{MI250}~\cite{amd-mi250-rocmdocs}, Ascend \texttt{910B}~\cite{liao2021ascend}), while \texttt{FP8} support remains limited to newer architectures.
Addintionally, we fine-tune \texttt{Qwen2.5-1.5B} and \texttt{Qwen2.5-3B} for 1,000 steps using either \texttt{INT8} or \texttt{FP8 SageBwd} (both with $\mathbf{dOV}^\top$ kept in \texttt{FP16} for fairness), and then inference with \texttt{FP4 SageAttention3}.
As shown in Table~\ref{tab:int8_fp8_finetune}, models fine-tuned with \texttt{INT8} attention achieve higher accuracy on both GSM8K and MMLU benchmarks.

\begin{table}[h]
\vspace{-.5em}
\centering
\small
\begin{minipage}{0.48\linewidth}
\centering
\caption{L1 error of $Q$, $K$, and $V$ gradients.}
\label{tab:int8_vs_fp8_l1}
\begin{tabular}{lccc}
\toprule
Method & $dQ$ $\downarrow$ & $dK$ $\downarrow$ & $dV$ $\downarrow$ \\
\midrule
\texttt{INT8 SageBwd} & \textbf{0.0290} & \textbf{0.0317} & \textbf{0.0423} \\
\texttt{FP8 SageBwd} & 0.0696 & 0.0999 & 0.0873 \\
\bottomrule
\end{tabular}
\end{minipage}
\hfill
\begin{minipage}{0.48\linewidth}
\centering
\caption{Cos similarity of $Q$, $K$, and $V$ gradients.}
\label{tab:int8_vs_fp8_cos}
\begin{tabular}{lccc}
\toprule
Method & $dQ$ $\uparrow$ & $dK$ $\uparrow$ & $dV$ $\uparrow$ \\
\midrule
\texttt{INT8 SageBwd} & \textbf{0.9987} & \textbf{0.9993} & \textbf{0.9995} \\
\texttt{FP8 SageBwd} & 0.9880 & 0.9910 & 0.9955 \\
\bottomrule
\end{tabular}
\end{minipage}
\vspace{-.5em}
\end{table}

\vspace{-5mm}
\begin{table}[h]
\centering
\small
\caption{Comparison of \texttt{INT8} and \texttt{FP8} \texttt{SageBwd} fine-tuning on \texttt{Qwen2.5} models.}
\label{tab:int8_fp8_finetune}
\vspace{-0.5em}
\begin{subtable}[t]{0.48\textwidth}
    \centering
    \subcaption{\texttt{Qwen2.5-1.5B}}
    \label{tab:qwen15b}
    \scalebox{0.95}{
    \begin{tabular}{lcc}
    \toprule
    Method & GSM8K $\uparrow$ & MMLU $\uparrow$ \\
    \midrule
    \texttt{INT8} Fine-tuning & \textbf{0.5232} & \textbf{0.4934} \\
    \texttt{FP8} Fine-tuning & 0.5031 & 0.4689 \\
    \bottomrule
    \end{tabular}}
\end{subtable}
\hfill
\begin{subtable}[t]{0.48\textwidth}
    \centering
    \small
    \subcaption{\texttt{Qwen2.5-3B}}
    \label{tab:qwen3b}
    \scalebox{0.95}{
    \begin{tabular}{lcc}
    \toprule
    Method & GSM8K $\uparrow$ & MMLU $\uparrow$ \\
    \midrule
    \texttt{INT8} Fine-tuning & \textbf{0.5945} & \textbf{0.6032} \\
    \texttt{FP8} Fine-tuning & 0.5868 & 0.5907 \\
    \bottomrule
    \end{tabular}}
\end{subtable}
\vspace{-.75em}
\end{table}

% \section{Discussion}
% Xxx.

\section{Related Work}
Recent efficient attention works~\citep{zhangsurvey} that utilize hardware features to accelerate attention computation methods mainly include the following: FlashAttention~\citep{dao2022flashattention} introduces tiling to reduce the GPU memory I/O between global memory and on-chip SRAM, achieving significant speedup. FlashAttention2~\citep{dao2023flashattention} improves the parallelism and warp partition strategies. FlashAttention3~\citep{shah2024flashattention} exclusively optimizes the kernel speed on the Hopper GPUs. xformers~\citep{xFormers2022} accelerates attention using dedicated CUDA kernels. SageAttention~\citep{2024sageattention} and SageAttention2~\citep{zhang2024sageattention2,zhang2025sageattention2++} accelerate attention using quantization and some novel outlier smoothing techniques. 
RingAttention~\citep{liuringattention} extends FlashAttention to multi-GPU/Node environments. In these works, although FlashAttention3 proposes a version of \texttt{FP8} attention, it has failed to be applied to video generation models in a plug-and-play way~\citep{zhang2024sageattention2}. Moreover, the \texttt{FP8} attention in FlashAttention3 does not support the backward pass, limiting its applicability to training tasks. Additionally, numerous efficient attention variants have emerged, including linear attention~\citep{wang2020linformer,choromanski2020rethinking,yu2022metaformer,katharopoulos2020transformers,qin2024lightning,yang2024gated} and sparse attention~\citep{zhang2025spargeattn,zhang2025sla,xi2025sparse,yang2025sparse,liu2021swin,chu2021twins,xiao2023efficient,xiao2024infllm,chen2023longlora,jiang2024minference,venkataramanan2023skip,gao2024seerattention,moaattention}. Although these works represent promising research directions, they are orthogonal to our work.

\section{Conclusions}
In this paper, we make two key contributions. Firstly, we design \our, the first microscaling \texttt{FP4} attention for inference acceleration, achieving \textbf{1038} \texttt{TOPS} on \texttt{RTX5090}, which is a \textbf{5}$\times$ speedup than the fastest FlashAttention on \texttt{RTX5090}. Experiments show that \our could accelerate various models with no end-to-end quality metrics degradation.
Secondly, we introduce the first trainable \texttt{8-bit} attention (\sageback) for training acceleration and explore its feasibility in training tasks. We find that the \texttt{8-bit} attention could achieve lossless performance in fine-tuning tasks, but currently has some limitations in pertaining tasks.

\textbf{Future Work.} First, while \sageback demonstrates faster performance than \texttt{FP16} implementation, we observe a noticeable gap between its current speed and theoretical upper bounds. This gap may be caused by suboptimal Triton kernel implementations, which we plan to further optimize. Second, and more importantly, investigating the application of low-bit attention in pretraining tasks presents a promising research direction worthy of exploration.

\section*{Acknowledgments}
This work was supported by the NSFC Projects (Nos. 62550004, 92270001, 62376131). J.Z is also supported by the XPlorer Prize.

% \newpage

%\bibliographystyle{neurips_2025}
%\bibliographystyle{IEEEtran}
\bibliographystyle{unsrt}
\bibliography{reference.bib}

\newpage
\appendix
\section{Appendix}

\subsection{Visible Comparison Examples}  \label{sec:appedix_visible_example}
\begin{figure}[H]
    \centering
    \includegraphics[width=.75\columnwidth]{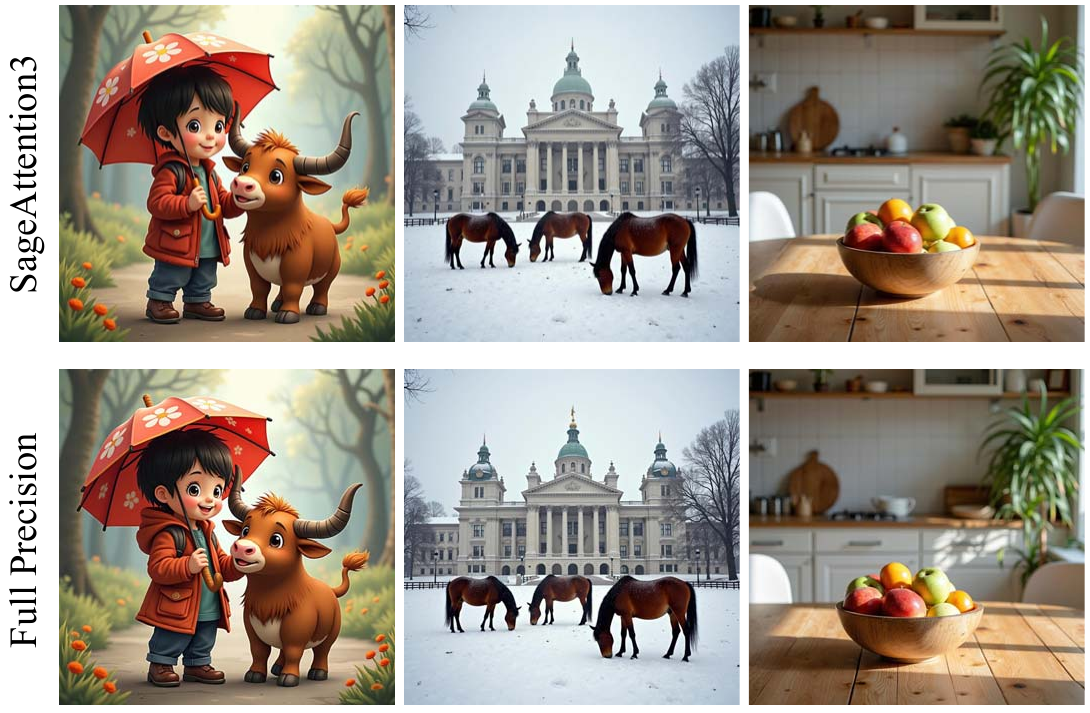} 
    \caption{Visible examples of image generation on \texttt{Stable-Diffusion3.5}.}
    \label{fig:sd_visual} 
\end{figure}

\begin{figure}[H]
    \centering
    \includegraphics[width=.75\columnwidth]{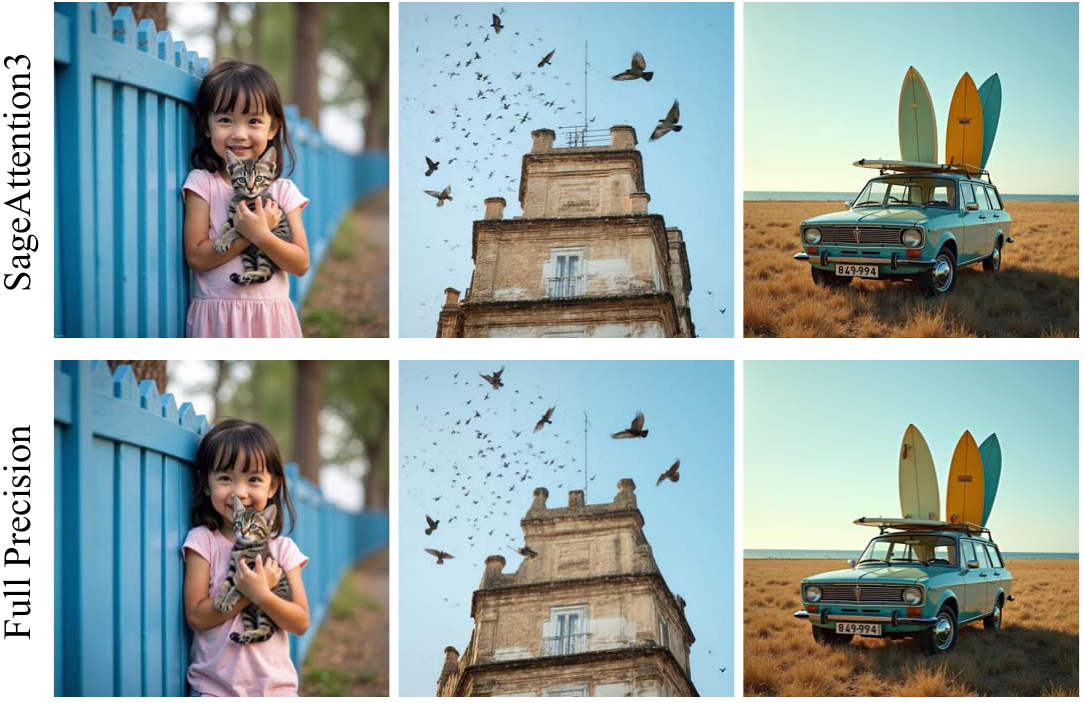} 
    \caption{Visible examples of image generation on \texttt{Flux}.}
    \label{fig:flux_visual} 
\end{figure}

\begin{figure}[H]
    \centering
    \begin{subfigure}[b]{0.3\textwidth}
        \includegraphics[width=\textwidth]{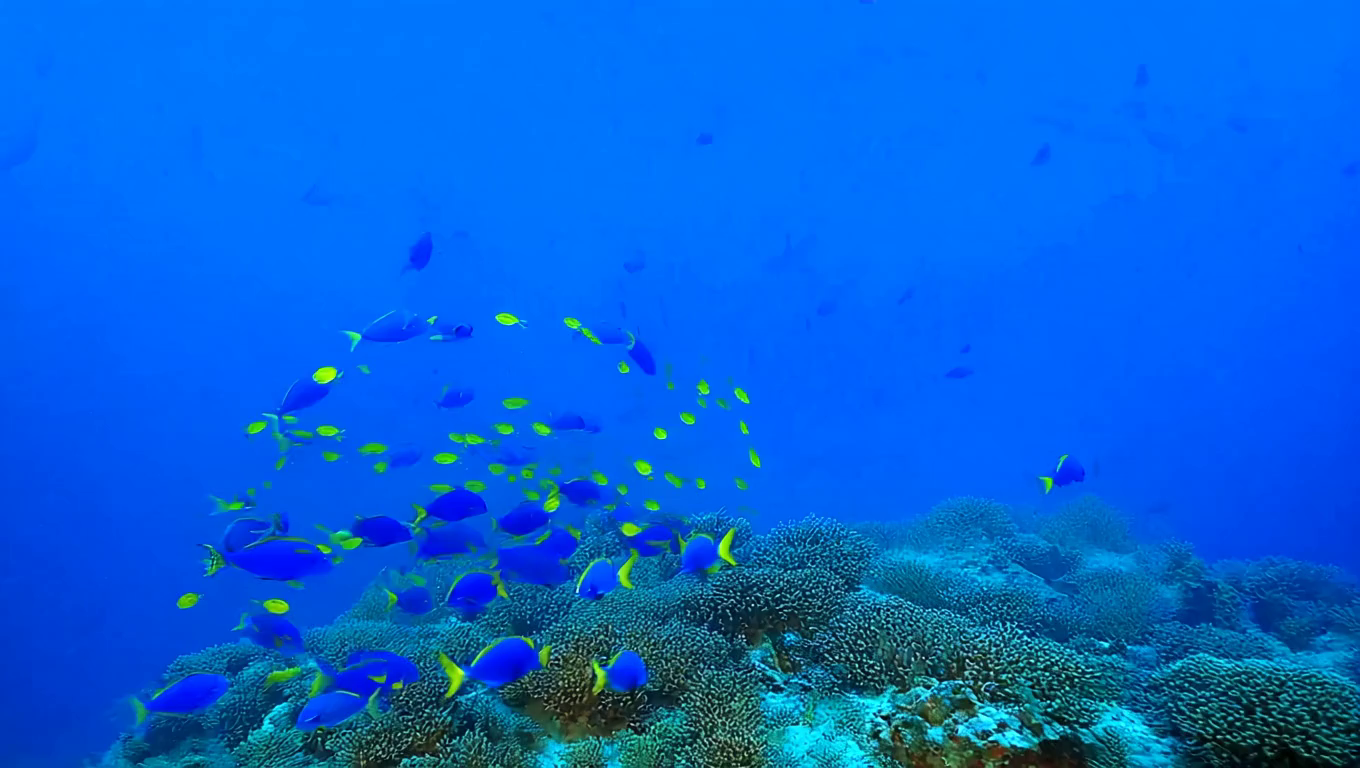}
        \caption{Full-Precision}
        \label{fig:a}
    \end{subfigure}
    \hfill
    \begin{subfigure}[b]{0.3\textwidth}
        \includegraphics[width=\textwidth]{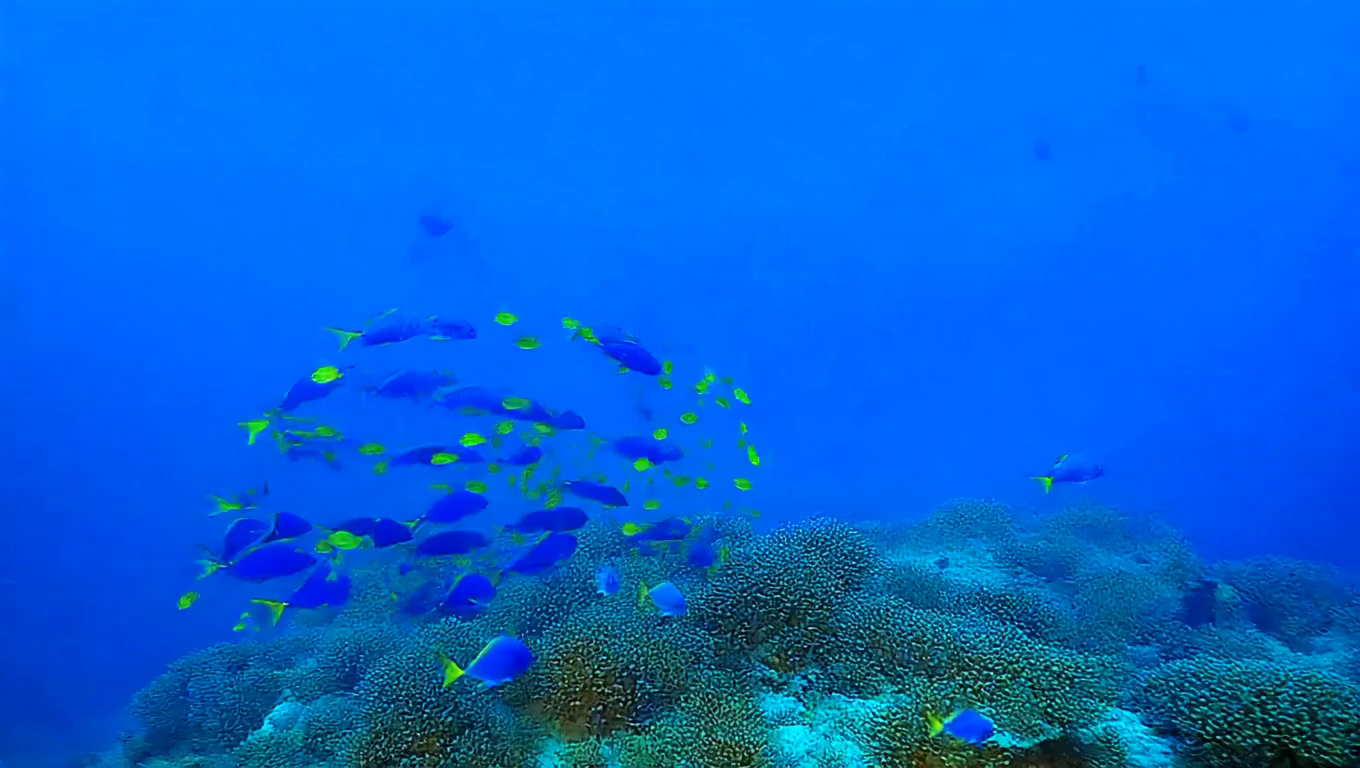}
        \caption{Two-level quantization}
        \label{fig:b}
    \end{subfigure}
    \hfill
    \begin{subfigure}[b]{0.3\textwidth}
        \includegraphics[width=\textwidth]{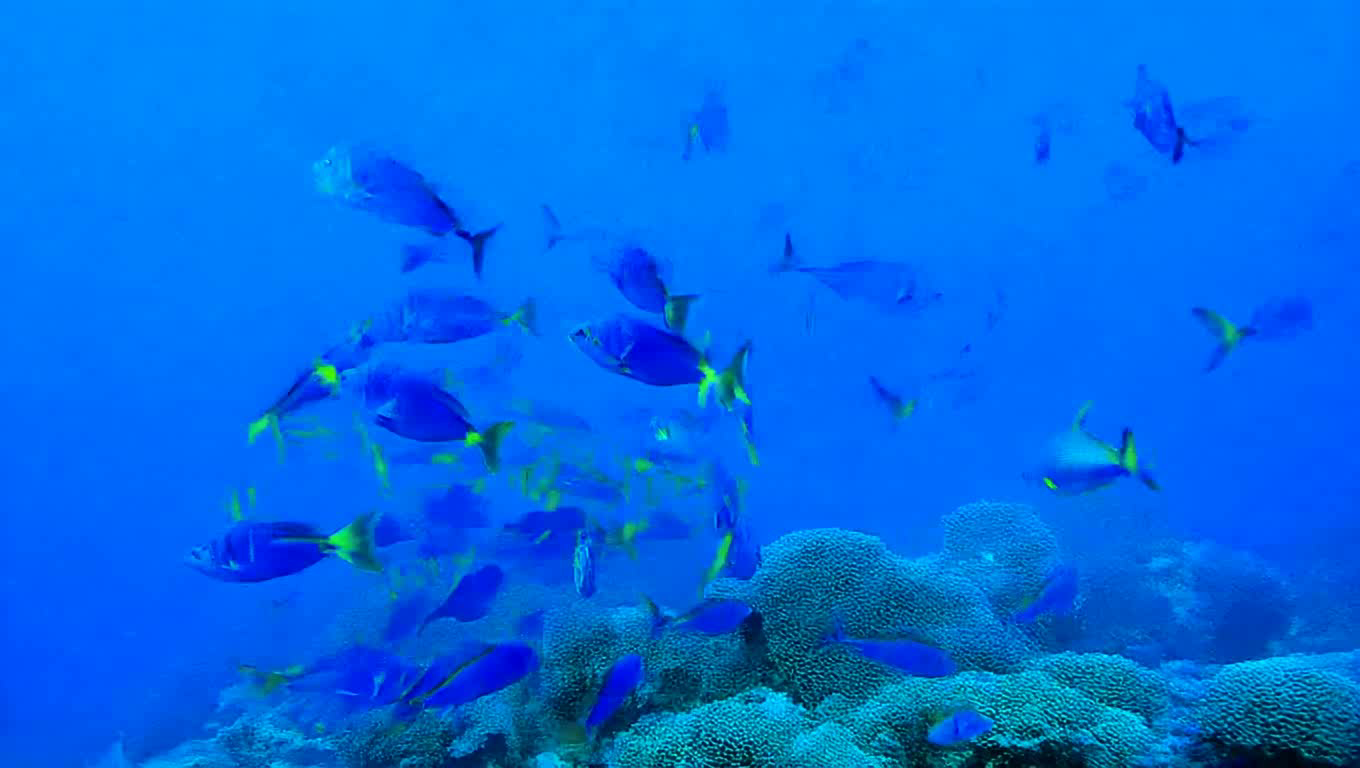}
        \caption{Direct quantization}
        \label{fig:c}
    \end{subfigure}
    \caption{Visual comparison of different scale strategies for $\widetilde \vP$ from \texttt{CogVideoX}.}
    \label{fig:scale_ablation}
\end{figure}

\begin{figure}[H]
    \centering
    \includegraphics[width=.885\columnwidth]{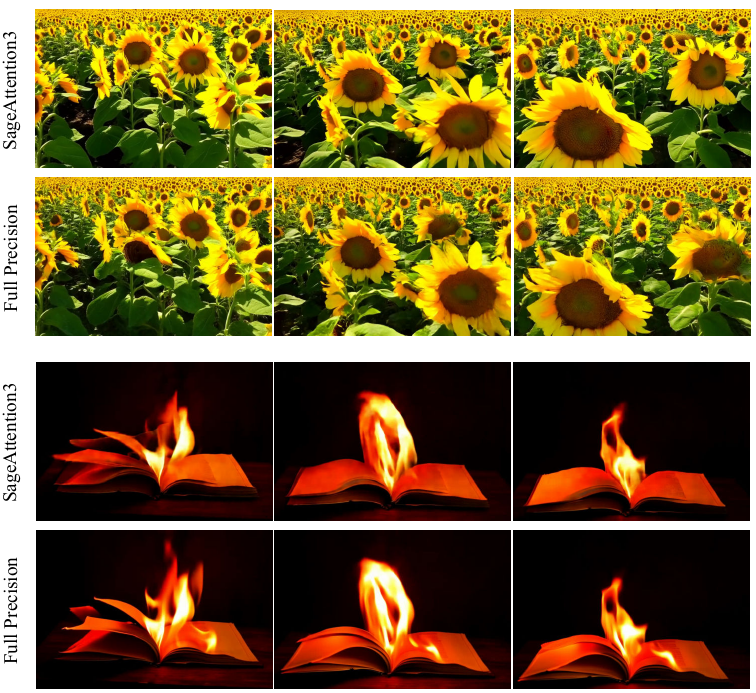} 
    \vspace{-.25em}
    \caption{Visible examples of video generation on \texttt{CogVideoX}.}
    \vspace{-1.5em}
    \label{fig:cog_visual} 
\end{figure}

\begin{figure}[H]
    \centering
    \includegraphics[width=.885\columnwidth]{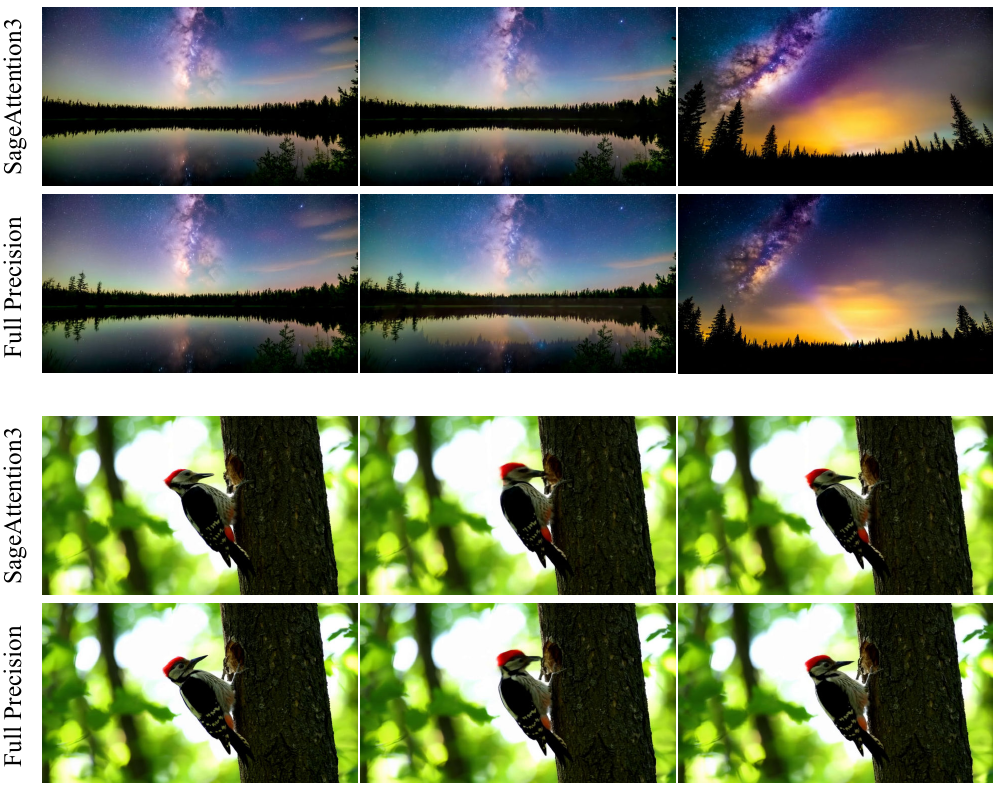} 
    \vspace{-.25em}
    \caption{Visible examples of video generation on \texttt{HunyuanVideo}.}
    \label{fig:hunyuan_visual} 
\end{figure}

Fig.~\ref{fig:sd_visual} and Fig.~\ref{fig:flux_visual} show additional visual comparison examples of image generation tasks. Fig.~\ref{fig:cog_visual} and Fig.~\ref{fig:hunyuan_visual} show more visual comparison examples of video generation tasks.

\subsection{Additional Kernel Speed Comparison}  \label{sec:appedix_kernel_speed}

\begin{figure}[H]
  \centering
  \includegraphics[width=\columnwidth]{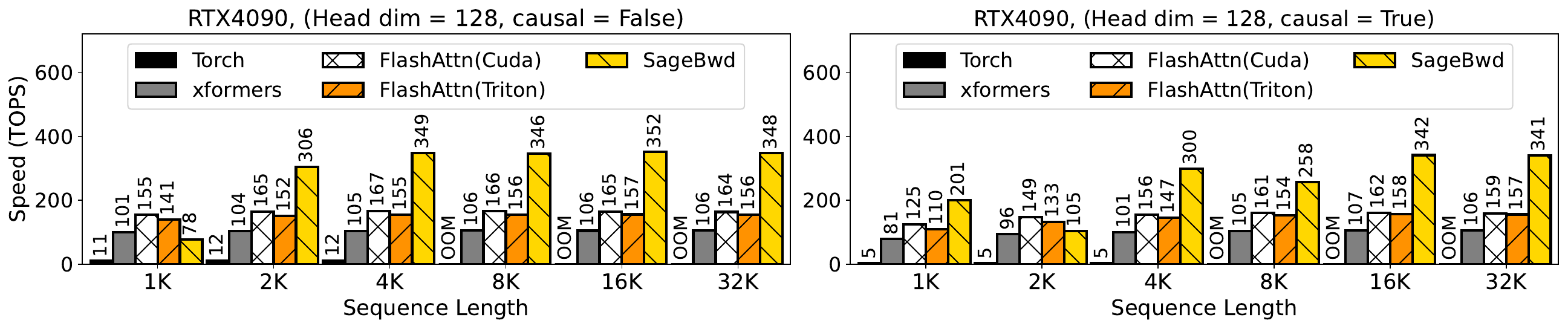} 
  \vspace{-1em}
  \caption{Forward speed comparison between \texttt{SageBwd} and Baselines (\texttt{RTX4090}, headim=128).}
  \label{fig:sage_train_fwd_h128_speed} 
\end{figure}

\begin{figure}[H]
  \centering
  \includegraphics[width=\columnwidth]{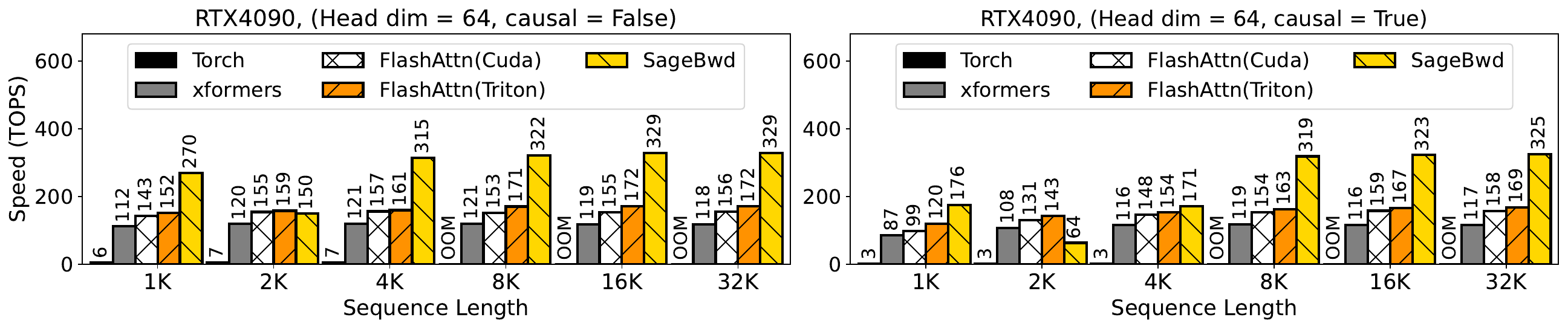} 
  \vspace{-1em}
  \caption{Forward speed comparison between \texttt{SageBwd} and Baselines (\texttt{RTX4090}, headim=64).}
  \label{fig:sage_train_fwd_h64_speed} 
\end{figure}

\begin{figure}[H]
  \centering
  \includegraphics[width=\columnwidth]{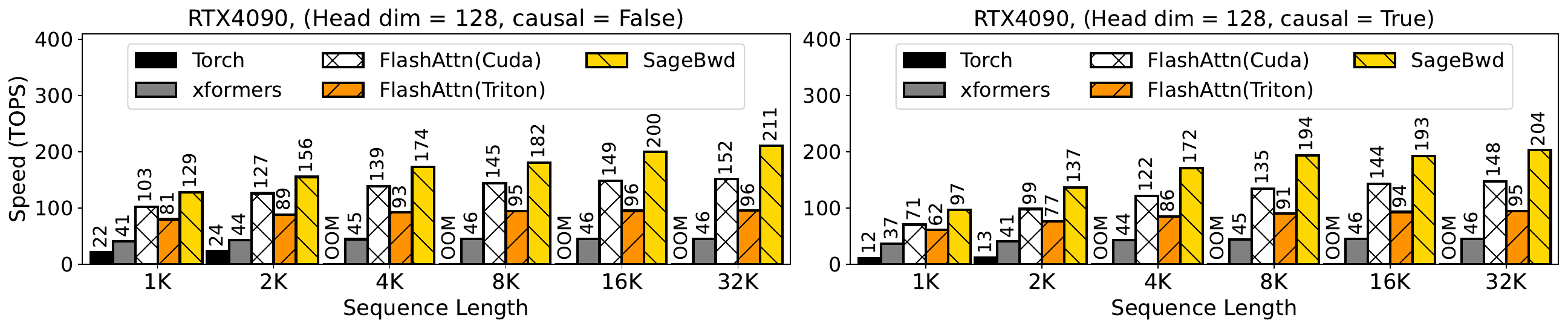} 
  \vspace{-1em}
  \caption{Backward speed comparison between \texttt{SageBwd} and Baselines (\texttt{RTX4090}, headim=128).}
  \label{fig:sage_train_bwd_h128_speed} 
\end{figure}

\begin{figure}[H]
    \centering
    \includegraphics[width=\columnwidth]{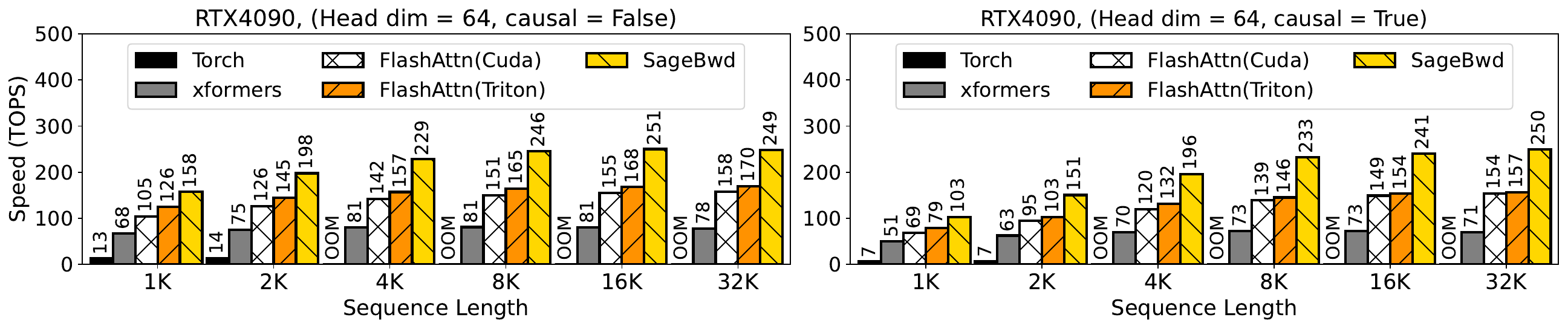} 
    \vspace{-1em}
    \caption{Backward speed comparison between \texttt{SageBwd} and Baselines (\texttt{RTX4090}, headim=64).}
    \label{fig:sage_train_bwd_h64_speed} 
\end{figure}

Fig.~\ref{fig:sage_train_fwd_h128_speed} and Fig.~\ref{fig:sage_train_fwd_h64_speed} show the forward kernel speed of \texttt{SageBwd}. Fig.~\ref{fig:sage_train_bwd_h128_speed} and Fig.~\ref{fig:sage_train_bwd_h64_speed} show the backward kernel speed of \texttt{SageBwd}. \texttt{SageBwd} achieved a \textbf{2x} speed up than FlashAttention in the forward propagation. \texttt{SageBwd} achieved a 1.2\textasciitilde\textbf{1.6x} speed up than FlashAttention in the backward propagation.


\subsection{Datasets, Metrics, and Hyperparameters}  \label{sec:exp_dataset_metrics}
\noindent \noindent\textbf{Datasets.}   
Text-to-video models are evaluated using the open-sora~\citep{opensora} prompt sets.
Text-to-image models are assessed on COCO annotations~\citep{lin2014microsoft}.
Language models are evaluated on GSM8K~\citep{cobbe2021gsm8k}, DROP~\citep{Dua2019DROP}, MMLU~\citep{MMLU}, and HELLASWAG~\citep{zellers2019hellaswag} datasets.

\noindent \noindent\textbf{End-to-end metrics.}   
For text-to-text models, we use Accuracy (Acc.) and F1-Score (F1).
For text-to-video models, we evaluate the quality of generated videos on five metrics: CLIPSIM and CLIP-Temp (CLIP-T)~\cite{liu2024evalcrafter} to measure the text-video alignment; (VQA-a) and (VQA-t) to assess the video aesthetic and technical quality, respectively; and Flow-score (FScore) for temporal consistency~\cite{wu2023exploring}. 
For text-to-image models, generated images are evaluated in three aspects: FID~\cite{heusel2017gans} and sFID~\cite{salimans2016improved} for fidelity evaluation, \textit{Clipscore} (CLIP)~\cite{hessel2021clipscore} for text-image alignment, and \textit{ImageReward} (IR)~\cite{xu2024imagereward} for human preference.

\noindent\textbf{Accuracy metrics.} We use three metrics to assess the accuracy of quantized attention output $O'$ compared to attention output in full-precision $O$: First, we flatten $O'$ and $O$ into vectors in the shape of $1\times n$. Then, Cosine similarity: $CosSim=\sum OO' / \smash{\sqrt{\sum O^2}} \smash{\sqrt{\sum O'^2}}$, Relative L1 distance: $L1=\sum |O - O'| / \sum |O|$, Root mean square error: $RMSE=\sqrt{(1/n) \sum (O - O')^2}$.

\textbf{Hyperparameters.} 
For pretraining tasks, we use a 400M model with a hidden size of 1024, 20 layers, an intermediate size of 3072, and 16 attention heads. The training uses a learning rate of 1e-3 with linear decay over 1000 warmup steps, and each step processes 2M tokens.
For finetuning tasks, we train for 700 steps using a learning rate of 3e-5 with linear decay and 100 warmup steps with a batch size of 32 on GSM8K dataset and 128 on MMLU, DROP, and HELLASWAG datasets.

\begin{figure}[H]
    \centering
    \includegraphics[width=1\columnwidth]{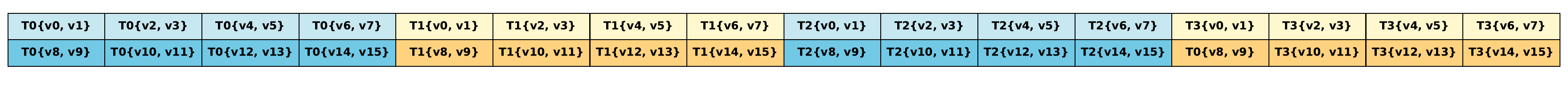} 
    \vspace{-1em}
    \caption{\texttt{FP4} operand A register layout - rows 0 and 8, thread 0-3, entries 0-15.}
    \label{fig:input_layout} 
\end{figure}
\begin{figure}[H]
    \centering
    \includegraphics[width=1\columnwidth]{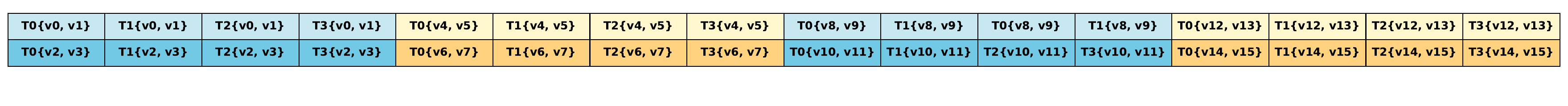} 
    \vspace{-1em}
    \caption{\texttt{FP32} accumulator register layout - rows 0 and 8, thread 0-3, entries 0-15.}
    \label{fig:ori_fp32_layout} 
\end{figure}
\begin{figure}[H]
    \centering
    \includegraphics[width=1\columnwidth]{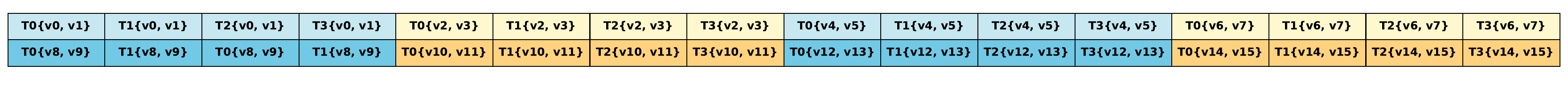} 
    \vspace{-1em}
    \caption{Permuted \texttt{FP32} accumulator register layout - rows 0 and 8, thread 0-3, entries 0-15.}
    \label{fig:new_fp32_layout} 
\end{figure}

\subsection{Additional Experiments of using \texttt{SageBwd}}  \label{sec:appendix_exp}
Table~\ref{tab:seed_1.5b_1}--\ref{tab:seed_llama_2} show \qwen(1.5B), \qwen(3B), and \llamal(3B) fine-tuning results on four datasets with five different random seeds. The average and standard deviation show \texttt{SageBwd} is highly consistent with \texttt{BF16} across various random seeds.

\begin{table}[H]
\centering
\caption{Comparison of \texttt{SageBwd} and \texttt{BF16} performance on GSM8K and DROP across different seeds on \qwen(1.5B).}
\label{tab:seed_1.5b_1}
\scalebox{0.98}{
\begin{tabular}{cccccc}
\toprule
\multirow{2}{*}{Seed} & \multicolumn{2}{c}{GSM8K} & \multicolumn{2}{c}{DROP} \\
\cmidrule(lr){2-3} \cmidrule(lr){4-5}
& \texttt{SageBwd} & \texttt{BF16} & \texttt{SageBwd} & \texttt{BF16} \\
\midrule
\num{42}   & \num{0.5133} & \num{0.5125} & \num{0.7316} & \num{0.7364} \\
\num{233}  & \num{0.5027} & \num{0.5042} & \num{0.7269} & \num{0.7295} \\
\num{1234} & \num{0.4973} & \num{0.4973} & \num{0.7329} & \num{0.7342} \\
\num{5678} & \num{0.5201} & \num{0.5208} & \num{0.7340} & \num{0.7332} \\
\num{1}    & \num{0.5049} & \num{0.5057} & \num{0.7278} & \num{0.7404} \\
\midrule
\textbf{Avg} & \num{0.5077} & \num{0.5081} & \num{0.7307} & \num{0.7348} \\
\textbf{Std} & \num{0.0090} & \num{0.0089} & \num{0.0032} & \num{0.0040} \\
\bottomrule
\end{tabular}}
\end{table}

\begin{table}[H]
\centering
\caption{Comparison of \texttt{SageBwd} and \texttt{BF16} performance on MMLU and HellaSwag across different seeds on \qwen(1.5B).}
\label{tab:seed_1.5b_2}
\scalebox{0.98}{
\begin{tabular}{cccccc}
\toprule
\multirow{2}{*}{Seed} & \multicolumn{2}{c}{MMLU} & \multicolumn{2}{c}{HellaSwag} \\
\cmidrule(lr){2-3} \cmidrule(lr){4-5}
& \texttt{SageBwd} & \texttt{BF16} & \texttt{SageBwd} & \texttt{BF16} \\
\midrule
\num{42}   & \num{0.5814} & \num{0.5873} & \num{0.9089} & \num{0.9065} \\
\num{233}  & \num{0.5746} & \num{0.5785} & \num{0.9082} & \num{0.9049} \\
\num{1234} & \num{0.5805} & \num{0.5836} & \num{0.9025} & \num{0.9047} \\
\num{5678} & \num{0.5736} & \num{0.5693} & \num{0.9112} & \num{0.9053} \\
\num{1}    & \num{0.5830} & \num{0.5823} & \num{0.9058} & \num{0.9075} \\
\midrule
\textbf{Avg} & \num{0.5786} & \num{0.5802} & \num{0.9073} & \num{0.9058} \\
\textbf{Std} & \num{0.0043} & \num{0.0069} & \num{0.0033} & \num{0.0012} \\
\bottomrule
\end{tabular}}
\end{table}

\begin{table}[H]
\centering
\caption{Comparison of \texttt{SageBwd} and \texttt{BF16} performance on GSM8K and DROP across different seeds on \qwen(3B).}
\label{tab:seed_3b_1}
\scalebox{0.98}{
\begin{tabular}{cccccc}
\toprule
\multirow{2}{*}{Seed} & \multicolumn{2}{c}{GSM8K} & \multicolumn{2}{c}{DROP} \\
\cmidrule(lr){2-3} \cmidrule(lr){4-5}
& \texttt{SageBwd} & \texttt{BF16} & \texttt{SageBwd} & \texttt{BF16} \\
\midrule
\num{42}   & \num{0.5982} & \num{0.6232} & \num{0.7800} & \num{0.7812} \\
\num{233}  & \num{0.5997} & \num{0.5974} & \num{0.7786} & \num{0.7812} \\
\num{1234} & \num{0.6156} & \num{0.6103} & \num{0.7786} & \num{0.7824} \\
\num{5678} & \num{0.6065} & \num{0.6012} & \num{0.7816} & \num{0.7853} \\
\num{1}    & \num{0.6171} & \num{0.6073} & \num{0.7813} & \num{0.7832} \\
\midrule
\textbf{Avg} & \num{0.6074} & \num{0.6079} & \num{0.7800} & \num{0.7827} \\
\textbf{Std} & \num{0.0001} & \num{0.0001} & \num{0.0000} & \num{0.0000} \\
\bottomrule
\end{tabular}}
\end{table}

\begin{table}[H]
\centering
\caption{Comparison of \texttt{SageBwd} and \texttt{BF16} performance on MMLU and HellaSwag across different seeds on \qwen(3B).}
\label{tab:seed_3b_2}
\scalebox{0.98}{
\begin{tabular}{cccccc}
\toprule
\multirow{2}{*}{Seed} & \multicolumn{2}{c}{MMLU} & \multicolumn{2}{c}{HellaSwag} \\
\cmidrule(lr){2-3} \cmidrule(lr){4-5}
& \texttt{SageBwd} & \texttt{BF16} & \texttt{SageBwd} & \texttt{BF16} \\
\midrule
\num{42}   & \num{0.6434} & \num{0.6425} & \num{0.9419} & \num{0.9402} \\
\num{233}  & \num{0.6431} & \num{0.6437} & \num{0.9405} & \num{0.9402} \\
\num{1234} & \num{0.6492} & \num{0.6492} & \num{0.9414} & \num{0.9429} \\
\num{5678} & \num{0.6531} & \num{0.6400} & \num{0.9430} & \num{0.9440} \\
\num{1}    & \num{0.6510} & \num{0.6454} & \num{0.9446} & \num{0.9434} \\
\midrule
\textbf{Avg} & \num{0.6480} & \num{0.6442} & \num{0.9423} & \num{0.9421} \\
\textbf{Std} & \num{0.0000} & \num{0.0000} & \num{0.0000} & \num{0.0000} \\
\bottomrule
\end{tabular}}
\end{table}

\begin{table}[H]
\centering
\caption{Comparison of \texttt{SageBwd} and \texttt{BF16} performance on GSM8K and DROP across different seeds on \llamal(1B).}
\label{tab:seed_llama_1}
\scalebox{0.98}{
\begin{tabular}{cccccc}
\toprule
\multirow{2}{*}{Seed} & \multicolumn{2}{c}{GSM8K} & \multicolumn{2}{c}{DROP} \\
\cmidrule(lr){2-3} \cmidrule(lr){4-5}
& \texttt{SageBwd} & \texttt{BF16} & \texttt{SageBwd} & \texttt{BF16} \\
\midrule
\num{42}   & \num{0.2722} & \num{0.2547} & \num{0.6367} & \num{0.6447} \\
\num{233}  & \num{0.2661} & \num{0.2570} & \num{0.6456} & \num{0.6424} \\
\num{1234} & \num{0.2616} & \num{0.2873} & \num{0.6439} & \num{0.6352} \\
\num{5678} & \num{0.2684} & \num{0.2585} & \num{0.6372} & \num{0.6409} \\
\num{1}    & \num{0.2646} & \num{0.2335} & \num{0.6393} & \num{0.6441} \\
\midrule
\textbf{Avg} & \num{0.2666} & \num{0.2582} & \num{0.6405} & \num{0.6414} \\
\textbf{Std} & \num{0.0000} & \num{0.0003} & \num{0.0000} & \num{0.0000} \\
\bottomrule
\end{tabular}}
\end{table}

\begin{table}[H]
\centering
\caption{Comparison of \texttt{SageBwd} and \texttt{BF16} performance on MMLU and HellaSwag across different seeds on \llamal(3B).}
\label{tab:seed_llama_2}
\scalebox{0.98}{
\begin{tabular}{cccccc}
\toprule
\multirow{2}{*}{Seed} & \multicolumn{2}{c}{MMLU} & \multicolumn{2}{c}{HellaSwag} \\
\cmidrule(lr){2-3} \cmidrule(lr){4-5}
& \texttt{SageBwd} & \texttt{BF16} & \texttt{SageBwd} & \texttt{BF16} \\
\midrule
\num{42}   & \num{0.4665} & \num{0.4705} & \num{0.8230} & \num{0.8319} \\
\num{233}  & \num{0.4646} & \num{0.4560} & \num{0.8327} & \num{0.8256} \\
\num{1234} & \num{0.4702} & \num{0.4757} & \num{0.8202} & \num{0.8243} \\
\num{5678} & \num{0.4580} & \num{0.4639} & \num{0.8232} & \num{0.8276} \\
\num{1}    & \num{0.4666} & \num{0.4691} & \num{0.8218} & \num{0.8236} \\
\midrule
\textbf{Avg} & \num{0.4652} & \num{0.4670} & \num{0.8242} & \num{0.8266} \\
\textbf{Std} & \num{0.0000} & \num{0.0000} & \num{0.0000} & \num{0.0000} \\
\bottomrule
\end{tabular}}
\end{table}

\subsection{Transposing $V$.}  \label{sec:appendix_others}

Performing the forward propagation of attention in full \texttt{FP4} precision poses additional challenges compared to \texttt{FP16}.
The input tensors $Q$, $K$, and $V$ are typically contiguous in the head dimensions. However, the row-major constraints on \texttt{FP4} MMA for the second GEMM necessitate $V$ to be contiguous in the sequence length dimension. Calling a standalone pre-processing transpose kernel for this purpose incurs excessive overhead, particularly during inference, which is often a memory-bound situation.
We address the problem by kernel fusion. For the first problem, we fuse the transpose of $V$ into the quantization kernel, thereby avoiding additional I/O overhead. 

\subsection{Accmulated Quantization Error Analysis.} 

\begin{table}[H]
\sisetup{detect-weight=true, detect-family=true}
\centering
\vspace{-.5em}
\caption{Layer-wise L1 error analysis of \texttt{SageAttention3} on \texttt{CogVideoX-2B}. 
The second row shows the results by retaining the three most sensitive layers in \texttt{FP16}.}
\label{tab:drift_analysis}
\scalebox{0.95}{
\begin{tabular}{@{}lcccc@{}}
\toprule
\textbf{Method} & \textbf{Layer1} $\downarrow$ & \textbf{Layer10} $\downarrow$ & \textbf{Layer20} $\downarrow$ & \textbf{Layer30} $\downarrow$ \\
\midrule
Use \texttt{SageAttention3} directly & 0.0076 & 0.0922 & 0.1146 & 0.0571 \\
Keep 3 most sensitive layers in \texttt{FP16} & 0.0076 & \textbf{0.0447} & \textbf{0.0773} & \textbf{0.0429} \\
\bottomrule
\end{tabular}}
\end{table}

To explore the issue of accumulated quantization error across layers, we conduct an analysis using \texttt{SageAttention3} on \texttt{CogVideoX-2B} and report the per-layer L1 error in Table \ref{tab:drift_analysis}. We observe that the accumulated error generally increases with layer depth, though it occasionally decreases in deeper layers, suggesting partial error cancellation.
To mitigate this drift, we apply a simple yet effective strategy: keeping the three layers with the largest observed error growth in \texttt{FP16} precision. As shown in the table, this adjustment significantly reduces the overall error accumulation across layers.

\subsection{Ablation of Smoothing Techniques.}

\begin{table}[H]
\sisetup{detect-weight=true, detect-family=true}
\centering
\vspace{-.5em}
\caption{Ablation of attention accuracy with different smoothing methods on \texttt{CogVideoX-2B}. 
Smoothing K and Smoothing Q are techniques from \texttt{SageAttention} and \texttt{SageAttention2}.
}
\label{tab:smooth_ablation}
\scalebox{0.95}{
\begin{tabular}{@{}lccc@{}}
\toprule
\textbf{Method} & \textbf{Cossim} $\uparrow$ & \textbf{L1 Error} $\downarrow$ & \textbf{RMSE} $\downarrow$ \\
\midrule
None & 0.915642 & 0.335867 & 0.303483 \\
SmoothQuant & 0.930125 & 0.267617 & 0.252883 \\
Hadamard & 0.941222 & 0.262047 & 0.223970 \\
\textbf{Smoothing\_Q} & \textbf{0.982848} & \textbf{0.115658} & \textbf{0.125862} \\
\textbf{Smoothing\_K} & \textbf{0.991176} & \textbf{0.094832} & \textbf{0.097668} \\
\bottomrule
\end{tabular}}
\end{table}

To investigate the impact of different smoothing strategies on attention accuracy, we compare several existing techniques, including SmoothQuant~\citep{xiao2023smoothquant} and Hadamard transformations, which provide per-token or per-tensor scaling control. However, we find these methods less effective in our setting.
\texttt{SageAttention3} inherits the smoothing Q and smoothing K mechanisms introduced in \texttt{SageAttention2}.
We conduct an ablation study on all layers of \texttt{CogVideoX-2B} to evaluate their effects.
As shown in Table~\ref{tab:smooth_ablation}, both smoothing Q and smoothing K yield substantially higher cosine similarity and lower reconstruction errors, demonstrating their effectiveness in stabilizing quantized attention computation.

\subsection{Theoretical Speed Comparison.} 

\begin{table}[H]
\sisetup{detect-weight=true, detect-family=true}
\centering
\vspace{-.5em}
\caption{Theoretical throughput comparison between FlashAttention3 and \texttt{SageAttention3} across different GPUs.}
\label{tab:theoretical_tops}
\scalebox{0.95}{
\begin{tabular}{@{}lccc@{}}
\toprule
\textbf{Method} & \textbf{B300 TOPS} $\uparrow$ & \textbf{B200 TOPS} $\uparrow$ & \textbf{RTX5090 TOPS} $\uparrow$ \\
\midrule
FlashAttention3 & 2500 & 2500 & 209.5 \\
FlashAttenion3 (\texttt{FP8}) & 5000 & 5000 & 419 \\
\textbf{\texttt{SageAttention3 (FP4)}} & \textbf{15000} & \textbf{10000} & \textbf{1676} \\
\bottomrule
\end{tabular}}
\end{table}

To provide a theoretical comparison with FlashAttention3, we refer to NVIDIA’s official documentation on throughput (TOPS) across different precisions. Since FlashAttention3 is currently only supported on \texttt{H100} GPUs, a direct empirical comparison is not feasible. Instead, we estimate the theoretical compute throughput of both FlashAttention3 and our \texttt{SageAttention3} on GPUs that support \texttt{FP4} Tensor Cores (\texttt{B300}, \texttt{B200}, and \texttt{RTX5090}).
As summarized in Table~\ref{tab:theoretical_tops}, \texttt{SageAttention3} achieves substantially higher theoretical peak throughput, highlighting its potential for further accelerating attention computation beyond FlashAttention-3.

\subsection{FlashAttentions vs SageAttentions.}

\begin{table}[H]
\sisetup{detect-weight=true, detect-family=true}
\centering
\vspace{-.5em}
\caption{Speed–accuracy trade-off of different attention methods.}
\label{tab:tradeoff}
\scalebox{0.95}{
\begin{tabular}{@{}lccc@{}}
\toprule
\textbf{Method} & \textbf{TOPS on 5090} $\uparrow$ & \textbf{TOPS on H100} $\uparrow$ & \textbf{Accuracy (CosSim)} $\uparrow$ \\
\midrule
FlashAttention2 & 214 & 338 & 100.000\% \\
FlashAttention3 (16bit) & N/A & 470 & 100.000\% \\
FlashAttention3 (8bit) & N/A & 890 & 98.570\% \\
SageAttention1 & 479 & 518 & 99.996\% \\
SageAttention2 (8bit) & 643 & 885 & 99.995\% \\
\texttt{SageAttention3 (4bit)} & \textbf{1038} & N/A & \textbf{99.551\%} \\
\bottomrule
\end{tabular}}
\label{tab:h100_5090_speed}
\end{table}

To illustrate the trade-off between accuracy and speed, we recorded the accuracy (Cosine similarity) of various attention methods across all layers of \texttt{CogVideoX-2B}, along with their theoretical throughput on \texttt{RTX5090} and \texttt{H100} GPUs. These results are summarized in the Table~\ref{tab:h100_5090_speed}.

% \begin{table}[h]
% \centering
% \caption{L1 Error of $Q$, $K$, and $V$ gradients.}
% \begin{tabular}{lccc}
% \toprule
% Method & $dQ$ L1 Error $\downarrow$ & $dK$ L1 Error $\downarrow$ & $dV$ L1 Error $\downarrow$ \\
% \midrule
% INT8 SageBwd & \textbf{0.0290} & \textbf{0.0317} & \textbf{0.0423} \\
% FP8 SageBwd & 0.0696 & 0.0999 & 0.0873 \\
% \bottomrule
% \end{tabular}
% \end{table}

% \begin{table}[h]
% \centering
% \caption{Cosine similarity of $Q$, $K$, and $V$ gradients.}
% \begin{tabular}{lccc}
% \toprule
% Method & $dQ$ CosSim $\uparrow$ & $dK$ CosSim $\uparrow$ & $dV$ CosSim $\uparrow$ \\
% \midrule
% INT8 SageBwd & \textbf{0.9987} & \textbf{0.9993} & \textbf{0.9995} \\
% FP8 SageBwd & 0.9880 & 0.9910 & 0.9955 \\
% \bottomrule
% \end{tabular}
% \end{table}

\subsection{Analysis of Two-Level Quantization.} 
\begin{proof}

We analyze the relative quantization error of $\widetilde \vP$ using both \texttt{direct quantization} and \texttt{two-level quantization} as follows:

For \texttt{direct quantization}, the relative quantization error, denoted as $E_1$, is defined as:
\begin{align}
    %\widetilde \vP =& \texttt{OnlineSoftmax}(\vS)  \notag \\
    \mathbf{s_P}, \hat \vP = \phi(\widetilde \vP), ~~~ E_1 = \frac{|\mathbf{s_P} \times \hat \vP - \widetilde \vP|}{|\widetilde \vP|}
    \label{err_1}
\end{align}

%(2) Two-level quantization:
%\begin{align*}
%    E_2 =  |\hat \vP_2 \times \mathbf{s_{P_2}} \times \mathbf{s_{P_1}} - \widetilde \vP|
%    = |\hat \vP_2 \times \mathbf{s_{P_2}}|
%\end{align*}

%The first-level quantization in two-level quantization is as follows:
For \texttt{two-level quantization}, the first level proceeds as:
\begin{align*}
    \mathbf{s_{P_1}} = \rowmax(\widetilde \vP) / (448 \times 6), ~~~
     \widetilde \vP_2 &= \widetilde \vP / \mathbf{s_{P_1}}, ~~~
\end{align*}

The quantization error introduced in this first step is negligible because $\widetilde \vP$, $\widetilde\vP_2$, and $s_{P_1}$ are all represented in \texttt{FP32} format.

We focus primarily on the second-level quantization, where the relative quantization error $E_2$ is given by:
\begin{align}
     \mathbf{s_{P_2}}, \hat \vP_2 = \phi (\widetilde \vP_2),~~~
     E_2 = \frac{|\mathbf{s_{p_2}} \times \hat \vP_2 - \widetilde \vP_2|}{|\widetilde \vP_2|}
     \label{err_2}
\end{align}

% The final error of two-level quantization is:
% \begin{align*}
%     E_{two-level} = E_2 / \mathbf{s_{P_1}}
% \end{align*}

The key difference between Equation~\ref{err_1} and \ref{err_2} lies in the range of the \texttt{FP8} scale factor.

Let $\{X\}_n$ denote the number of distinct representable values in the set $X$. Then:

in \texttt{direct quantization}:
\begin{align*}
    0 \le \mathbf{s_{P}} \le 0.167,~~~ \mathbf{s_{P}} \in \mathbf{E4M3}, ~~~ \{\mathbf{s_{P}}\}_n = 35
\end{align*}

In \texttt{two-level quantization}:
\begin{align*}
    0 \le \mathbf{s_{P_2}} \le 448.0, ~~~ \mathbf{s_{P_2}} \in \mathbf{E4M3}, ~~~ \{\mathbf{s_{P_2}}\}_n = 127
\end{align*}

Since $\{\textbf{E2M1}\}_n = 8$, the number of unique outputs after dequantization is:

For \texttt{direct quantization}:
\begin{align*}
    \widetilde \vP^{\prime} = \mathbf{s_{P}} \times \hat \vP, ~~~ \{\widetilde \vP^{\prime} \}_n = 35 \times 8 = 280
\end{align*}

For \texttt{two-level quantization}:
\begin{align*}
    \widetilde \vP_{2}^{\prime} = \mathbf{s_{P_2}} \times \hat \vP_2, ~~~ \{\widetilde \vP_2^{\prime} \}_n = 127 \times 8 = 1016
\end{align*}

Let $\Delta(p_i)$ denote the interval between the two nearest quantization levels surrounding the value $p_i \in \widetilde \vP$.
Then the absolute quantization error satisfies:
%interval between two neighbour quantization candidate values, We have:
\begin{align*}
|\hat p_i - p_i| \le \frac{\Delta(p_i)}{2}
\end{align*}
The relative error $\varepsilon$ satisfies:
\begin{align*}
   \varepsilon_i \le \frac{\Delta(p_i)}{2 \times p_i}
\end{align*}

Given that $\{\widetilde \vP_2^{\prime} \}_n > \{\widetilde \vP^{\prime} \}_n$, the quantization intervals in the two-level scheme are finer:
\begin{align*}
\frac{\Delta(\widetilde \vP_2)}{\widetilde \vP_2} < \frac{\Delta(\widetilde \vP)}{\widetilde \vP}
\end{align*}

%Thus, we have the upper bound relative error of two quantization strategy:
Thus, the relative quantization error satisfies:
\begin{align*}
\frac{|\widetilde \vP_2^{\prime} - \widetilde \vP_2|}{\widetilde \vP_2} <  \frac{|\vP^{\prime} - \widetilde \vP|}{\widetilde \vP} 
\end{align*}

Which leads to the conclusion:
\begin{align*}
    E_2 < E_1
\end{align*}
%Finished the proof.

\end{proof}

% To make the $E_1$ in Equation \ref{err_1} and Equation \ref{err_2} in the same magnitude, we can do as follows:
% \begin{align*}
%     E_1^{*} = |\mathbf{s_P} \times \hat \vP - \widetilde \vP| \times \frac{1}{\mathbf{s_{P_1}}}
% \end{align*}

%the scale factor $s_{P}$ ranging from 0 to about 0.167 in E4M3 format,

\subsection{Analysis of the Benefit of Keeping $\vdO_i\vV_j^\top$ in FP16 in \texttt{SageBwd}.}

The backward pass of \texttt{SageBwd} involves 5 MatMuls. The accuracy of $\vS_{ij}=\vQ_i\vK_j^\top$ is fully addressed in SageAttention2. The remaining four are as follows:

\begin{enumerate}
    \item[(1)] $\vdP_{ij}=\vdO_i\vV_j^\top$.
    \item[(2)] $\vdQ_i\gets\vdQ_i+\vdS_{ij}\vK_j$
    \item[(3)] $\vdK_j\gets\vdK_j+\vdS_{ij}^\top\vQ_i$
    \item[(4)] $\vdV_j\gets\vdV_j+\vP_{ij}^\top\vdO_i$
\end{enumerate}

We choose to keep (1) in FP16, while quantizing others to INT8. This choice can be formally justified:

\begin{proof}
Following~\citep{lin2023awq}, we assume that any matrix ${\bf X}\in\mathbb R^{n\times d}$ (e.g. $\vQ,\vK,\vV,\vdO$) satisfies:

\begin{itemize}
    \item The entries in $\bf X$ are mutually independent.
    \item ${\bf X}_{ij} \sim N(\mu_{{\bf X},j},\sigma_{{\bf X},j}^2)$, i.e. the distribution of each token is identical.
\end{itemize}

The quantization error of a matrix $\bf X$ is denoted as:
$$
\Delta{\bf X}:=s_{\bf X}\hat{\bf X} - {\bf X},\quad \text{where }s_{\bf X},\hat{\bf X} = \psi({\bf X}).
$$

For example, consider the error in $\vdQ$. Neglecting second-order error terms, we have:
$$
\Delta\vdQ = \underbrace{(\vP \circ (\vdO\Delta\vV^\top + \Delta \vdO\vV^\top))\vK}_{\Delta\vdQ^{(1)} \text{ from (1)}} + \underbrace{\Delta \vdS \vK + \vdS \Delta \vK}_{\vdQ^{(2)}\text{from (2)}}.
$$

Here, $\vdS=\vP\circ(\vdP-D)=\vP\circ(\vdO\vV^\top-D)$, where $D=\vdO\odot\vO$. In element-wise terms (the subscript denotes a single element):
$$
\vdS_{ij} = \vP_{ij} \sum_k \vdO_{ik} (\vV_{jk} - \vO_{ik}) = \vP_{ij}\sum_k \vdO_{ik}\left(\vV_{jk} - \sum_\ell \vP_{i\ell}\vV_{\ell k}\right)
$$

Since $\vV$ is independent of other variables, by linearity of expectation:
$$
\mathbb E[\vdS_{ij}] = \mathbb E\left[\vP_{ij}\sum_k\vdO_{ik}\left(\mu_{\vV,k}-\sum_\ell\vP_{i\ell} \mu_{\vV,k}\right)\right] = 0.
$$

Moreover, as negating $\vV$ flips the sign of $\vdS_{ij}$, the PDF of $\vdS_{ij}$ is symmetric. Using a "round-to-nearest" quantization policy, we have $\mathbb E[\Delta\vdS]=0$. Thus
$$
\mathbb E\left[\vdQ^{(2)}\right] = \mathbb E[\Delta\vdS\vK + \vdS\Delta\vK] = 0,
$$

while $\mathbb E\left[\Delta\vdQ^{(1)}\right]$ is generally non-zero (e.g. when distributions have non-zero means), indicating that $\vdQ$'s error is dominated by $\Delta\vdQ^{(1)}$.

\end{proof}

\subsection{Broader Impact}
This paper presents work that aims to advance the field of efficient machine learning systems. It can be used to accelerate the inference and training processes of various models. None of the negative impacts we feel must be specifically highlighted here.

\clearpage
\newpage
\section*{NeurIPS Paper Checklist}

\begin{enumerate}

\item {\bf Claims}
    \item[] Question: Do the main claims made in the abstract and introduction accurately reflect the paper's contributions and scope?
    \item[] Answer: \answerYes{} % Replace by \answerYes{}, \answerNo{}, or \answerNA{}.
    \item[] Justification: Both the Abstract and the Introduction indicate the scope and contributions of this paper.
    \item[] Guidelines:
    \begin{itemize}
        \item The answer NA means that the abstract and introduction do not include the claims made in the paper.
        \item The abstract and/or introduction should clearly state the claims made, including the contributions made in the paper and important assumptions and limitations. A No or NA answer to this question will not be perceived well by the reviewers. 
        \item The claims made should match theoretical and experimental results, and reflect how much the results can be expected to generalize to other settings. 
        \item It is fine to include aspirational goals as motivation as long as it is clear that these goals are not attained by the paper. 
    \end{itemize}

\item {\bf Limitations}
    \item[] Question: Does the paper discuss the limitations of the work performed by the authors?
    \item[] Answer: \answerYes{} % Replace by \answerYes{}, \answerNo{}, or \answerNA{}.
    \item[] Justification: Limitations are discussed in Introduction and Conclusion.
    \item[] Guidelines:
    \begin{itemize}
        \item The answer NA means that the paper has no limitation while the answer No means that the paper has limitations, but those are not discussed in the paper. 
        \item The authors are encouraged to create a separate "Limitations" section in their paper.
        \item The paper should point out any strong assumptions and how robust the results are to violations of these assumptions (e.g., independence assumptions, noiseless settings, model well-specification, asymptotic approximations only holding locally). The authors should reflect on how these assumptions might be violated in practice and what the implications would be.
        \item The authors should reflect on the scope of the claims made, e.g., if the approach was only tested on a few datasets or with a few runs. In general, empirical results often depend on implicit assumptions, which should be articulated.
        \item The authors should reflect on the factors that influence the performance of the approach. For example, a facial recognition algorithm may perform poorly when image resolution is low or images are taken in low lighting. Or a speech-to-text system might not be used reliably to provide closed captions for online lectures because it fails to handle technical jargon.
        \item The authors should discuss the computational efficiency of the proposed algorithms and how they scale with dataset size.
        \item If applicable, the authors should discuss possible limitations of their approach to address problems of privacy and fairness.
        \item While the authors might fear that complete honesty about limitations might be used by reviewers as grounds for rejection, a worse outcome might be that reviewers discover limitations that aren't acknowledged in the paper. The authors should use their best judgment and recognize that individual actions in favor of transparency play an important role in developing norms that preserve the integrity of the community. Reviewers will be specifically instructed to not penalize honesty concerning limitations.
    \end{itemize}

\item {\bf Theory assumptions and proofs}
    \item[] Question: For each theoretical result, does the paper provide the full set of assumptions and a complete (and correct) proof?
    \item[] Answer: \answerYes{} % Replace by \answerYes{}, \answerNo{}, or \answerNA{}.
    \item[] Justification: We provide and check the proof for theoretical results.
    \item[] Guidelines:
    \begin{itemize}
        \item The answer NA means that the paper does not include theoretical results. 
        \item All the theorems, formulas, and proofs in the paper should be numbered and cross-referenced.
        \item All assumptions should be clearly stated or referenced in the statement of any theorems.
        \item The proofs can either appear in the main paper or the supplemental material, but if they appear in the supplemental material, the authors are encouraged to provide a short proof sketch to provide intuition. 
        \item Inversely, any informal proof provided in the core of the paper should be complemented by formal proofs provided in appendix or supplemental material.
        \item Theorems and Lemmas that the proof relies upon should be properly referenced. 
    \end{itemize}

    \item {\bf Experimental result reproducibility}
    \item[] Question: Does the paper fully disclose all the information needed to reproduce the main experimental results of the paper to the extent that it affects the main claims and/or conclusions of the paper (regardless of whether the code and data are provided or not)?
    \item[] Answer: \answerYes{} % Replace by \answerYes{}, \answerNo{}, or \answerNA{}.
    \item[] Justification: Result can be reproduced according to the sections of Method, Experiment, and Appendix~\ref{sec:exp_dataset_metrics} sections.
    \item[] Guidelines:
    \begin{itemize}
        \item The answer NA means that the paper does not include experiments.
        \item If the paper includes experiments, a No answer to this question will not be perceived well by the reviewers: Making the paper reproducible is important, regardless of whether the code and data are provided or not.
        \item If the contribution is a dataset and/or model, the authors should describe the steps taken to make their results reproducible or verifiable. 
        \item Depending on the contribution, reproducibility can be accomplished in various ways. For example, if the contribution is a novel architecture, describing the architecture fully might suffice, or if the contribution is a specific model and empirical evaluation, it may be necessary to either make it possible for others to replicate the model with the same dataset, or provide access to the model. In general. releasing code and data is often one good way to accomplish this, but reproducibility can also be provided via detailed instructions for how to replicate the results, access to a hosted model (e.g., in the case of a large language model), releasing of a model checkpoint, or other means that are appropriate to the research performed.
        \item While NeurIPS does not require releasing code, the conference does require all submissions to provide some reasonable avenue for reproducibility, which may depend on the nature of the contribution. For example
        \begin{enumerate}
            \item If the contribution is primarily a new algorithm, the paper should make it clear how to reproduce that algorithm.
            \item If the contribution is primarily a new model architecture, the paper should describe the architecture clearly and fully.
            \item If the contribution is a new model (e.g., a large language model), then there should either be a way to access this model for reproducing the results or a way to reproduce the model (e.g., with an open-source dataset or instructions for how to construct the dataset).
            \item We recognize that reproducibility may be tricky in some cases, in which case authors are welcome to describe the particular way they provide for reproducibility. In the case of closed-source models, it may be that access to the model is limited in some way (e.g., to registered users), but it should be possible for other researchers to have some path to reproducing or verifying the results.
        \end{enumerate}
    \end{itemize}

\item {\bf Open access to data and code}
    \item[] Question: Does the paper provide open access to the data and code, with sufficient instructions to faithfully reproduce the main experimental results, as described in supplemental material?
    \item[] Answer: \answerYes{} % Replace by \answerYes{}, \answerNo{}, or \answerNA{}.
    \item[] Justification: Source code and instructions will be provided in supplementary materials.
    \item[] Guidelines:
    \begin{itemize}
        \item The answer NA means that paper does not include experiments requiring code.
        \item Please see the NeurIPS code and data submission guidelines (\url{https://nips.cc/public/guides/CodeSubmissionPolicy}) for more details.
        \item While we encourage the release of code and data, we understand that this might not be possible, so “No” is an acceptable answer. Papers cannot be rejected simply for not including code, unless this is central to the contribution (e.g., for a new open-source benchmark).
        \item The instructions should contain the exact command and environment needed to run to reproduce the results. See the NeurIPS code and data submission guidelines (\url{https://nips.cc/public/guides/CodeSubmissionPolicy}) for more details.
        \item The authors should provide instructions on data access and preparation, including how to access the raw data, preprocessed data, intermediate data, and generated data, etc.
        \item The authors should provide scripts to reproduce all experimental results for the new proposed method and baselines. If only a subset of experiments are reproducible, they should state which ones are omitted from the script and why.
        \item At submission time, to preserve anonymity, the authors should release anonymized versions (if applicable).
        \item Providing as much information as possible in supplemental material (appended to the paper) is recommended, but including URLs to data and code is permitted.
    \end{itemize}

\item {\bf Experimental setting/details}
    \item[] Question: Does the paper specify all the training and test details (e.g., data splits, hyperparameters, how they were chosen, type of optimizer, etc.) necessary to understand the results?
    \item[] Answer: \answerYes{} % Replace by \answerYes{}, \answerNo{}, or \answerNA{}.
    \item[] Justification: Please refer to the Sections of Experiment and Appendex~\ref{sec:exp_dataset_metrics}.
    \item[] Guidelines:
    \begin{itemize}
        \item The answer NA means that the paper does not include experiments.
        \item The experimental setting should be presented in the core of the paper to a level of detail that is necessary to appreciate the results and make sense of them.
        \item The full details can be provided either with the code, in appendix, or as supplemental material.
    \end{itemize}

\item {\bf Experiment statistical significance}
    \item[] Question: Does the paper report error bars suitably and correctly defined or other appropriate information about the statistical significance of the experiments?
    \item[] Answer: \answerYes{} % Replace by \answerYes{}, \answerNo{}, or \answerNA{}.
    \item[] Justification: As listed in the Appendix, for those experiments with large errors, we provide mean and standard deviation values of the results.
    \item[] Guidelines:
    \begin{itemize}
        \item The answer NA means that the paper does not include experiments.
        \item The authors should answer "Yes" if the results are accompanied by error bars, confidence intervals, or statistical significance tests, at least for the experiments that support the main claims of the paper.
        \item The factors of variability that the error bars are capturing should be clearly stated (for example, train/test split, initialization, random drawing of some parameter, or overall run with given experimental conditions).
        \item The method for calculating the error bars should be explained (closed form formula, call to a library function, bootstrap, etc.)
        \item The assumptions made should be given (e.g., Normally distributed errors).
        \item It should be clear whether the error bar is the standard deviation or the standard error of the mean.
        \item It is OK to report 1-sigma error bars, but one should state it. The authors should preferably report a 2-sigma error bar than state that they have a 96\% CI, if the hypothesis of Normality of errors is not verified.
        \item For asymmetric distributions, the authors should be careful not to show in tables or figures symmetric error bars that would yield results that are out of range (e.g. negative error rates).
        \item If error bars are reported in tables or plots, The authors should explain in the text how they were calculated and reference the corresponding figures or tables in the text.
    \end{itemize}

\item {\bf Experiments compute resources}
    \item[] Question: For each experiment, does the paper provide sufficient information on the computer resources (type of compute workers, memory, time of execution) needed to reproduce the experiments?
    \item[] Answer: \answerYes{} % Replace by \answerYes{}, \answerNo{}, or \answerNA{}.
    \item[] Justification: We provide the specific computer resources.
    \item[] Guidelines:
    \begin{itemize}
        \item The answer NA means that the paper does not include experiments.
        \item The paper should indicate the type of compute workers CPU or GPU, internal cluster, or cloud provider, including relevant memory and storage.
        \item The paper should provide the amount of compute required for each of the individual experimental runs as well as estimate the total compute. 
        \item The paper should disclose whether the full research project required more compute than the experiments reported in the paper (e.g., preliminary or failed experiments that didn't make it into the paper). 
    \end{itemize}
    
\item {\bf Code of ethics}
    \item[] Question: Does the research conducted in the paper conform, in every respect, with the NeurIPS Code of Ethics \url{https://neurips.cc/public/EthicsGuidelines}?
    \item[] Answer: \answerYes{} % Replace by \answerYes{}, \answerNo{}, or \answerNA{}.
    \item[] Justification: The authors have reviewed the NeurIPS Code of Ethics.
    \item[] Guidelines:
    \begin{itemize}
        \item The answer NA means that the authors have not reviewed the NeurIPS Code of Ethics.
        \item If the authors answer No, they should explain the special circumstances that require a deviation from the Code of Ethics.
        \item The authors should make sure to preserve anonymity (e.g., if there is a special consideration due to laws or regulations in their jurisdiction).
    \end{itemize}

\item {\bf Broader impacts}
    \item[] Question: Does the paper discuss both potential positive societal impacts and negative societal impacts of the work performed?
    \item[] Answer: \answerYes{} % Replace by \answerYes{}, \answerNo{}, or \answerNA{}.
    \item[] Justification: Please refer to Appendix.
    \item[] Guidelines:
    \begin{itemize}
        \item The answer NA means that there is no societal impact of the work performed.
        \item If the authors answer NA or No, they should explain why their work has no societal impact or why the paper does not address societal impact.
        \item Examples of negative societal impacts include potential malicious or unintended uses (e.g., disinformation, generating fake profiles, surveillance), fairness considerations (e.g., deployment of technologies that could make decisions that unfairly impact specific groups), privacy considerations, and security considerations.
        \item The conference expects that many papers will be foundational research and not tied to particular applications, let alone deployments. However, if there is a direct path to any negative applications, the authors should point it out. For example, it is legitimate to point out that an improvement in the quality of generative models could be used to generate deepfakes for disinformation. On the other hand, it is not needed to point out that a generic algorithm for optimizing neural networks could enable people to train models that generate Deepfakes faster.
        \item The authors should consider possible harms that could arise when the technology is being used as intended and functioning correctly, harms that could arise when the technology is being used as intended but gives incorrect results, and harms following from (intentional or unintentional) misuse of the technology.
        \item If there are negative societal impacts, the authors could also discuss possible mitigation strategies (e.g., gated release of models, providing defenses in addition to attacks, mechanisms for monitoring misuse, mechanisms to monitor how a system learns from feedback over time, improving the efficiency and accessibility of ML).
    \end{itemize}
    
\item {\bf Safeguards}
    \item[] Question: Does the paper describe safeguards that have been put in place for responsible release of data or models that have a high risk for misuse (e.g., pretrained language models, image generators, or scraped datasets)?
    \item[] Answer: \answerNA{} % Replace by \answerYes{}, \answerNo{}, or \answerNA{}.
    \item[] Justification: The paper does not release models, generators, or datasets with a high risk for misuse.
    \item[] Guidelines:
    \begin{itemize}
        \item The answer NA means that the paper poses no such risks.
        \item Released models that have a high risk for misuse or dual-use should be released with necessary safeguards to allow for controlled use of the model, for example by requiring that users adhere to usage guidelines or restrictions to access the model or implementing safety filters. 
        \item Datasets that have been scraped from the Internet could pose safety risks. The authors should describe how they avoided releasing unsafe images.
        \item We recognize that providing effective safeguards is challenging, and many papers do not require this, but we encourage authors to take this into account and make a best faith effort.
    \end{itemize}

\item {\bf Licenses for existing assets}
    \item[] Question: Are the creators or original owners of assets (e.g., code, data, models), used in the paper, properly credited and are the license and terms of use explicitly mentioned and properly respected?
    \item[] Answer: \answerYes{} % Replace by \answerYes{}, \answerNo{}, or \answerNA{}.
    \item[] Justification: Please refer to the Experiment Section and the Appendix.
    \item[] Guidelines:
    \begin{itemize}
        \item The answer NA means that the paper does not use existing assets.
        \item The authors should cite the original paper that produced the code package or dataset.
        \item The authors should state which version of the asset is used and, if possible, include a URL.
        \item The name of the license (e.g., CC-BY 4.0) should be included for each asset.
        \item For scraped data from a particular source (e.g., website), the copyright and terms of service of that source should be provided.
        \item If assets are released, the license, copyright information, and terms of use in the package should be provided. For popular datasets, \url{paperswithcode.com/datasets} has curated licenses for some datasets. Their licensing guide can help determine the license of a dataset.
        \item For existing datasets that are re-packaged, both the original license and the license of the derived asset (if it has changed) should be provided.
        \item If this information is not available online, the authors are encouraged to reach out to the asset's creators.
    \end{itemize}

\item {\bf New assets}
    \item[] Question: Are new assets introduced in the paper well documented and is the documentation provided alongside the assets?
    \item[] Answer: \answerYes{} % Replace by \answerYes{}, \answerNo{}, or \answerNA{}.
    \item[] Justification: Please refer to README.md in our following supplemental material submission.
    \item[] Guidelines: 
    \begin{itemize}
        \item The answer NA means that the paper does not release new assets.
        \item Researchers should communicate the details of the dataset/code/model as part of their submissions via structured templates. This includes details about training, license, limitations, etc. 
        \item The paper should discuss whether and how consent was obtained from people whose asset is used.
        \item At submission time, remember to anonymize your assets (if applicable). You can either create an anonymized URL or include an anonymized zip file.
    \end{itemize}

\item {\bf Crowdsourcing and research with human subjects}
    \item[] Question: For crowdsourcing experiments and research with human subjects, does the paper include the full text of instructions given to participants and screenshots, if applicable, as well as details about compensation (if any)? 
    \item[] Answer: \answerNA{} % Replace by \answerYes{}, \answerNo{}, or \answerNA{}.
    \item[] Justification: The paper does not involve crowdsourcing nor research with human subjects. 
    \item[] Guidelines:
    \begin{itemize}
        \item The answer NA means that the paper does not involve crowdsourcing nor research with human subjects.
        \item Including this information in the supplemental material is fine, but if the main contribution of the paper involves human subjects, then as much detail as possible should be included in the main paper. 
        \item According to the NeurIPS Code of Ethics, workers involved in data collection, curation, or other labor should be paid at least the minimum wage in the country of the data collector. 
    \end{itemize}

\item {\bf Institutional review board (IRB) approvals or equivalent for research with human subjects}
    \item[] Question: Does the paper describe potential risks incurred by study participants, whether such risks were disclosed to the subjects, and whether Institutional Review Board (IRB) approvals (or an equivalent approval/review based on the requirements of your country or institution) were obtained?
    \item[] Answer: \answerNA{} % Replace by \answerYes{}, \answerNo{}, or \answerNA{}.
    \item[] Justification: The paper does not involve crowdsourcing nor research with human subjects.
    \item[] Guidelines:
    \begin{itemize}
        \item The answer NA means that the paper does not involve crowdsourcing nor research with human subjects.
        \item Depending on the country in which research is conducted, IRB approval (or equivalent) may be required for any human subjects research. If you obtained IRB approval, you should clearly state this in the paper. 
        \item We recognize that the procedures for this may vary significantly between institutions and locations, and we expect authors to adhere to the NeurIPS Code of Ethics and the guidelines for their institution. 
        \item For initial submissions, do not include any information that would break anonymity (if applicable), such as the institution conducting the review.
    \end{itemize}

\item {\bf Declaration of LLM usage}
    \item[] Question: Does the paper describe the usage of LLMs if it is an important, original, or non-standard component of the core methods in this research? Note that if the LLM is used only for writing, editing, or formatting purposes and does not impact the core methodology, scientific rigorousness, or originality of the research, declaration is not required.
    %this research? 
    \item[] Answer: \answerNA{} % Replace by \answerYes{}, \answerNo{}, or \answerNA{}.
    \item[] Justification: The core method development in this research does not involve LLMs as any important, original, or non-standard components.
    \item[] Guidelines:
    \begin{itemize}
        \item The answer NA means that the core method development in this research does not involve LLMs as any important, original, or non-standard components.
        \item Please refer to our LLM policy (\url{https://neurips.cc/Conferences/2025/LLM}) for what should or should not be described.
    \end{itemize}

\end{enumerate}
\end{document}